\documentclass{article}

\usepackage[preprint]{neurips_2026}

\usepackage[utf8]{inputenc} 
\usepackage[T1]{fontenc}    
\usepackage{hyperref}       
\usepackage{url}            
\usepackage{booktabs}       
\usepackage{amsfonts}       
\usepackage{nicefrac}       
\usepackage{microtype}      
\usepackage{xcolor}         
\usepackage[table]{xcolor}
\definecolor{baselinecolor}{RGB}{235,235,235}
\setcitestyle{numbers,square}
\usepackage{makecell}
\usepackage{tcolorbox}

\title{Generating the Wild: Individual-Consistent Image-to-Video Generation for Wildlife}

\usepackage{amsmath}
\usepackage{amssymb}
\usepackage{mathtools}
\usepackage{bm}
\usepackage{graphicx}
\usepackage{subcaption}
\usepackage{multirow}
\usepackage{enumitem}
\usepackage{comment}

\author{
Yuzhuo Li\thanks{Corresponding author.}
\quad
Di Zhao
\quad
Xinyu Zhang
\quad
Daniel Wilson
\quad
Yun Sing Koh
\\
School of Computer Science, University of Auckland\\
Auckland, New Zealand\\
\texttt{yil708@aucklanduni.ac.nz}\\
\texttt{\{di.zhao,xinyu.zhang,daniel.wilson,y.koh\}@auckland.ac.nz}
}

\begin{document}

\maketitle

\begin{abstract}
Individual-level wildlife identification often suffers from data scarcity, as varying observations of the same animal under diverse poses, viewpoints, and motions are rarely available. 
Image-to-video (I2V) generation offers a promising way to mitigate this limitation by synthesizing additional observations from a single reference image.
However, existing I2V models mainly emphasize global layout, semantics, and motion, and therefore often fail to preserve fine-grained local appearance cues that distinguish one wildlife individual from another, such as fur texture, stripe boundaries, spot configurations, and contour transitions.
We observe that these identity-critical cues are closely related to high-frequency information, which is essential for distinguishing individuals.
To address this challenge, we propose \textbf{WildIcon}, a high-frequency-guided I2V framework for wildlife individual consistency. 
Specifically, WildIcon introduces a frequency-aware identity encoding branch that extracts individual-specific high-frequency cues from the reference image.
Combined with isolated foreground information, the resulting identity tokens are then injected into cross-attention blocks as identity conditioning.
Building on a frozen backbone with lightweight identity adaptation, WildIcon preserves fine-grained identity cues visible in the reference image while retaining the motion controllability and semantic fidelity of the base I2V model.
In addition, to support the training and evaluation of wildlife individual-consistent I2V, we construct \textbf{WildlifeVid}, a wildlife-centric video dataset with high-quality, temporally coherent clips and individual-level identity labels. 
Experiments on I2V generation and downstream animal re-identification (ReID) show that WildIcon achieves stronger individual consistency than existing baselines, and that its filtered outputs can serve as useful candidate training augmentations for downstream ReID.
\end{abstract}

\section{Introduction}
\label{sec:intro}
Wildlife conservation is essential for maintaining ecological balance, protecting endangered habitats, and sustaining essential ecosystem services~\citep{huang2018linking,wu2026overcoming}.
A key requirement in these efforts is individual-level identification for monitoring population dynamics, movement patterns, and individual condition in a scalable and non-invasive way~\citep{beery2023wild,xu2024advanced}.
However, reliable wildlife identification remains fundamentally constrained by data scarcity, since repeated observations of the same individual under diverse poses, viewpoints, and environmental conditions are often unavailable~\citep{schneider2019past,wu2026region,li2025metawild}.

To mitigate this limitation, prior work has explored data enrichment strategies, including data augmentation techniques (\textit{e.g.,} cropping, flipping, and color jittering)~\citep{shorten2019survey,chai2026text} and synthetic image generation~\citep{rombach2022high}.
While conventional data augmentations can improve model robustness against moderate changes in scale, illumination and color, they remain limited to transformations of already observed samples and cannot synthesize genuinely new poses or complex out-of-plane viewpoint changes~\citep{li2020deep,zhao2025balancing,qiao2026zero}. 
Image generation models can produce more diverse appearances and poses, but they operate on isolated frames without explicit temporal or geometric constraints, and therefore often fail to preserve subject identity under large viewpoint changes~\citep{zhu2023human,jakab2018unsupervised,song2026boosting,zhang2025let}.
Image-to-video (I2V) generation is therefore a more promising direction: given a reference image, it can synthesize additional observations of the same individual under new motions, poses, and views~\citep{kong2024hunyuanvideo, rong2026distillation, li2026let}. 

\begin{figure}[t]
    \centering
    \includegraphics[width=0.95\columnwidth]{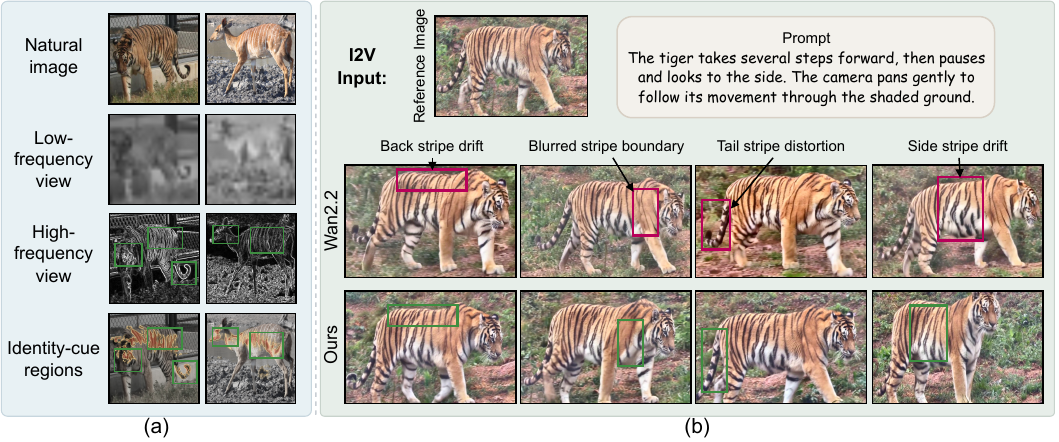}
    \caption{\textbf{Frequency-aware identity preservation in wildlife I2V generation.}
    (a) Low-frequency views preserve coarse body structure but suppress fine-grained identity cues, whereas high-frequency views reveal local identity-discriminative details.
    (b) Given the same reference image and prompt, Wan2.2 blurs or alters identity-critical stripe patterns across frames, while WildIcon better preserves these fine-grained cues for same-individual consistency.}
    \label{fig:intro_teaser}
\end{figure}

Despite this promise, existing I2V models are not designed for individual consistency in wildlife.
Although they can generate realistic motion and preserve coarse subject appearance, maintaining consistency with the reference animal across frames requires preserving subtle local appearance cues that remain discriminative under changes in pose, viewpoint, and background.
In wildlife, these cues often take the form of fine-grained visual patterns, such as 
fur or scale texture, stripe boundaries, spot configurations, and contour transitions~\citep{wu2026region,li2024adaptive}.
To better understand which visual information is most important for distinguishing individuals, we analyze the role of frequency components in wildlife identity. 
As shown in Figure~\ref{fig:intro_teaser},
these fine-grained patterns are more closely related to high-frequency image information, which captures rapid spatial changes and fine visual structures~\citep{lin2023deep,lee2026frequencybalanced}.
In contrast, low-frequency views mainly retain global content while suppressing many identity-critical details.
This observation reveals a mismatch between the requirements of wildlife identity preservation and the modeling priorities of modern I2V models: while they are effective at modeling global layout, semantic content and temporal motion, they do not explicitly preserve the sparse local high-frequency details critical for wildlife identity~\citep{yuan2025identity}. 
As a result, they may generate visually plausible videos while still drifting in identity-critical appearance from one individual to another.

Motivated by this observation, we propose \textbf{WildIcon} (Sec.~\ref{sec:method}), a high-frequency-guided I2V framework tailored to wildlife individual consistency. WildIcon introduces a frequency-aware identity encoding branch that explicitly extracts foreground-filtered high-frequency cues from the reference image and encodes them together with foreground appearance information into identity tokens for identity-aware conditioning.
These identity tokens are injected into selected cross-attention layers, enabling I2V models to preserve fine-grained individual identity, while the remaining blocks are kept frozen to retain the inherent motion controllability and semantic fidelity of the pretrained backbone.
To enable training and evaluation, we further construct \textbf{WildlifeVid}, a wildlife-centric video dataset curated through a scalable pipeline (Sec.~\ref{sec:dataset}) that provides high-quality, temporally coherent clips and individual identity labels under realistic variation in viewpoint, motion, illumination and background. Together, WildIcon and WildlifeVid provide a modeling and data foundation for wildlife individual-consistent I2V generation. 

We validate this framework on both wildlife I2V generation and downstream animal ReID.
Experiments show that WildIcon outperforms existing baselines
in individual consistency 
and downstream animal ReID performance by producing reliable, identity-aware synthetic videos.
{In summary, our main contributions} are:
(1) We construct WildlifeVid, a wildlife-centric video dataset comprising single-subject wildlife clips with identity labels and standardized prompts for training and evaluating individual-consistent I2V models.
(2) We propose WildIcon, a high-frequency-guided I2V framework that makes sparse wildlife identity patterns explicit conditioning signals, enabling
I2V models to preserve individual-specific 
local cues while maintaining motion controllability and semantic fidelity.
(3) We demonstrate that wildlife individual-consistent I2V benefits downstream animal ReID: by converting limited real observations into additional individual-consistent training samples, WildIcon helps mitigate the data scarcity that limits individual-level wildlife identification.

\section{Related Work}
\label{sec:related}

\textbf{Image-to-video generation.} 
Image-to-video (I2V) generation extends video foundation models by conditioning generation on a reference image and, optionally, a text prompt. 
Recent methods have explored a variety of conditioning strategies to inject visual guidance into pretrained video generators. 
For example, DynamiCrafter~\citep{xing2024dynamicrafter} incorporates image embeddings through cross-attention, while Stable Video Diffusion~\citep{blattmann2023stable} and CogVideoX~\citep{yang2025cogvideox} concatenate conditional latents with noisy video latents. 
Wan~\citep{wan2025wan} further adopts mask-guided conditioning and decoupled cross-attention to support high-quality I2V generation. 
While these methods improve controllability over text-only generation, maintaining fine-grained individual consistency remains challenging. 
For wildlife, identity cues are often sparse and closely tied to high-frequency structures, coarse reference-image conditioning is insufficient for individual consistency.
Although DiT-based I2V backbones model global structure and motion well, they lack explicit mechanisms for preserving local identity-critical details, leading to appearance drift in generated wildlife videos.

\textbf{Identity-preserving video generation.} 
Beyond general I2V generation, recent work has studied subject-consistent and identity-preserving video generation, where the goal is to maintain a reference subject across generated frames while following text instructions.
Subject-driven methods improve consistency through reference-image injection or cross-modal text-image-video alignment~\citep{wei2024dreamvideo,jiang2024videobooth,wu2025customcrafter,liu2025phantom}, while human-centered identity-preserving methods use facial representations, landmarks, or frequency-decomposed facial cues to preserve person identity~\citep{yuan2025identity,zhang2025magicmirror,tu2025stableanimator}.
Their conditioning mechanisms are mainly designed for general subject appearance or human facial identity, and therefore do not directly transfer to wildlife, where identity is often determined by sparse high-frequency patterns such as texture, stripes, spots, and contour transitions.
WildIcon instead conditions generation on foreground-filtered high-frequency cues to better maintain fine-grained identity-relevant patterns visible in the reference.

\textbf{Wildlife individual identity modeling.}
High-frequency information captures rapid spatial changes and fine visual structures, which in wildlife often correspond to individual-discriminative cues such as fur or scale texture, stripe boundaries, spot configurations, and contour details. 
AdaFreq~\citep{li2024adaptive} supports this view by showing that high-frequency cues provide an effective cross-species representation for wildlife ReID. 
While some recent generative efforts (\textit{e.g.}, Syn-Tiger-360~\citep{song2026boosting}) further show that preserving cross-view biometric topologies improves downstream ReID, generalizable identity-preserving I2V generation for diverse wildlife remains underexplored.
WildIcon extends wildlife identity modeling to this generative setting by using foreground-filtered high-frequency cues as conditioning signals for maintaining reference-supported individual appearance over time.

\section{WildlifeVid Dataset}
\label{sec:dataset}
\begin{figure*}[t]
    \centering
    \includegraphics[width=\textwidth]{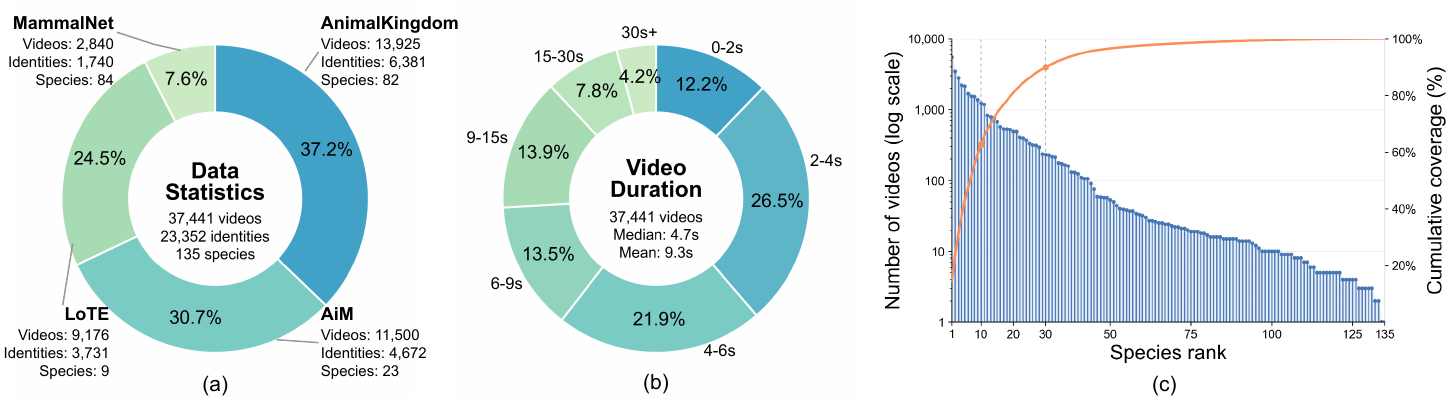}
    \caption{\textbf{Overview statistics of WildlifeVid.} (a) Composition of WildlifeVid across the four public data sources, showing the number of videos, identities, and species in each source. (b) Distribution of video durations. (c) Species distribution ranked by the number of videos, with cumulative coverage.}
    \label{fig:data-statistics}
\end{figure*}

Developing wildlife individual-consistent I2V models requires wildlife videos with individual-level identity annotations, because these labels indicate which clips correspond to the same wildlife individual and support learning the visual cues that distinguish it from others of the same species.
However, existing wildlife video datasets are primarily annotated for species, action, or behavior recognition, providing limited support for individual identity modeling.
To address this gap, we present WildlifeVid, a wildlife-centric video dataset that provides individual identity labels, curated via a scalable pipeline for wildlife-consistent I2V generation.
WildlifeVid is constructed by curating and re-annotating clips from four publicly accessible wildlife video sources: Animal-in-Motion (AiM)~\citep{zhao2025web}, AnimalKingdom~\citep{Ng_2022_CVPR}, MammalNet~\citep{chen2023mammalnet}, and LoTE-Animal~\citep{Liu_2023_ICCV}.
Figure~\ref{fig:data-statistics} summarizes the dataset statistics.

\subsection{Data Curation Pipeline}
\label{ssec:data-pipeline}
Since identity labels are assigned at the clip level, each retained clip should contain one dominant wildlife individual without identity changes over time.
We therefore curate raw videos into single-animal temporal clips, retaining complete videos when possible and splitting or filtering videos only when identity changes.
Each retained clip is paired with a standardized prompt and an individual identity label for individual-consistent I2V training.

\noindent\textbf{Animal-centric clip extraction.}
Raw wildlife videos often contain shot changes, frames without a visible animal, or multiple animals in the same view, making them unsuitable for identity annotation.
We therefore first split each source video into candidate clips based on shot changes using PySceneDetect~\citep{pyscenedetect}, which detects large weighted average pixel changes across adjacent frames. 
We then remove clips that do not clearly show the target species using CLIP-based species filtering~\citep{radford2021learning,hessel2021clipscore}.
Finally, we enforce a strict single-subject constraint on the remaining clips by retaining only clips with exactly one detected animal throughout the clip and discarding clips with zero or multiple detected animals.

\noindent\textbf{Visibility and temporal stability filtering.}
After obtaining candidate single-subject clips, we further filter them to ensure that each clip can be assigned a reliable individual identity label.
We sample frames from each candidate clip and examine the detected wildlife individual over time.
We remove clips in which the animal is missing, the bounding box shows large changes in location or scale across adjacent frames, or the visible individual may change within the clip.
Finally, we discard clips in which the wildlife's body is only partially visible within the frame, since incomplete bodies provide unreliable shape and appearance cues for identity learning.

\noindent\textbf{Prompt generation.}
Text prompts provide controllable conditions for wildlife I2V training, guiding the model to follow the desired motion, viewpoint, and scene context.
For each retained clip, we sample representative frames and use a multimodal captioning model to generate an initial caption.
We then convert the raw caption into a standardized prompt by removing generic template text (\textit{e.g.}, framing or style phrases that do not describe the visible content), unsupported inferences (\textit{e.g.}, guessed behavior, intent, or attributes not clearly visible), repeated phrases, and content unrelated to the visible animal or motion.
The final prompt (\textit{e.g.}, ``a tiger walks slowly through grass in a side view'') is a concise sentence describing the visible content of the clip.

\noindent\textbf{Identity annotation.}
We assign identity labels at the clip level by treating each retained clip as a candidate identity unit.
For each clip, we use the temporal midpoint frame as its representative frame and extract a visual embedding using DINOv3~\citep{simeoni2025dinov3}.
Candidate clips are grouped within the same species based on embedding similarity, which reduces cross-species mismatches and makes manual verification more tractable.
High-similarity candidates are then manually verified, and clips confirmed to contain the same animal are assigned the same identity label.

\begin{figure}[t]
    \centering
    \includegraphics[width=1.0\columnwidth]{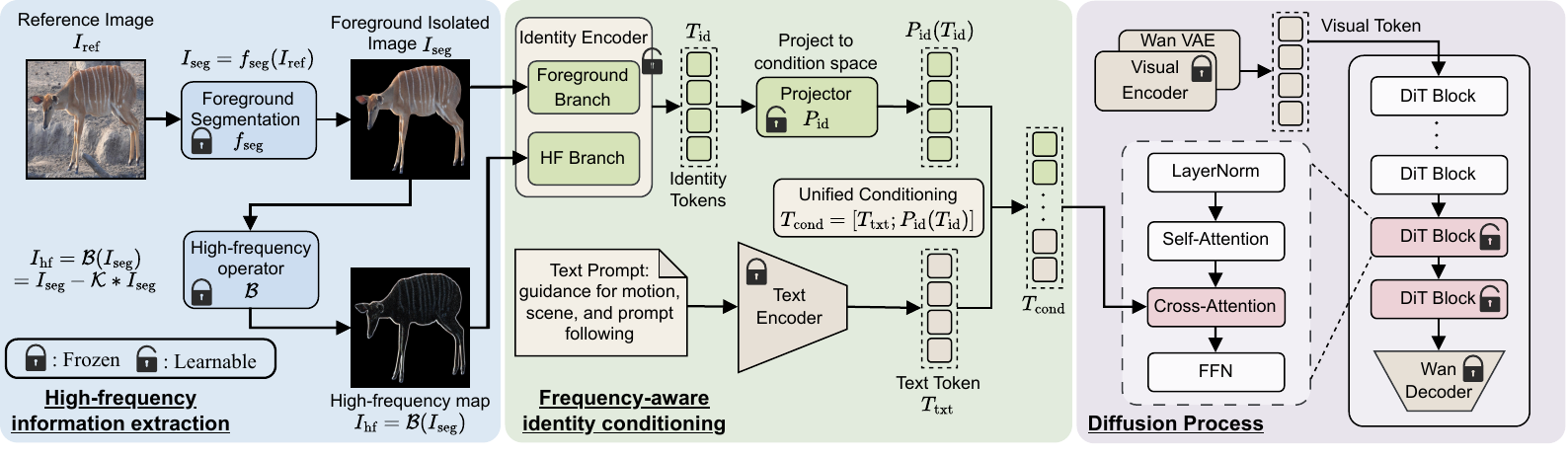}
    \caption{WildIcon augments a frozen Wan I2V backbone with a foreground-filtered high-frequency identity branch. Given a reference image $I_{\mathrm{ref}}$, the model first obtains a foreground-isolated image $I_{\mathrm{seg}}$ and its high-frequency map $I_{\mathrm{hf}}=\mathcal{B}(I_{\mathrm{seg}})$. These complementary cues are encoded into identity tokens $T_{\mathrm{id}}$, projected and fused with text tokens to form a unified condition sequence, and injected into selected middle and late DiT blocks through native cross-attention.} 
    \label{fig:framework}
\end{figure}

\section{Methodology}
\label{sec:method}
In this section, we propose \textbf{WildIcon}, a high-frequency-guided I2V framework for individual-consistent wildlife generation.
Given a reference image $I_{\mathrm{ref}}$ of a known wildlife individual and a text prompt $y$ describing the desired motion or scene context, our goal is to generate a video $V=\{I_t\}_{t=1}^{T}$ consisting of $T$ frames that remains temporally coherent, follows the prompt, and preserves the individual's fine-grained identity throughout the sequence. 
As illustrated in Figure~\ref{fig:framework}, WildIcon consists of a frequency-aware identity encoder, an identity-conditioning module, and a DiT-based video backbone.
Built on Wan2.2, WildIcon strengthens identity preservation by extracting foreground-filtered high-frequency cues from the reference image and injecting the resulting identity tokens into selected DiT blocks.

\subsection{Preliminary}
\label{sec:preliminary}
Our method is built on Wan2.2-I2V-A14B~\citep{wan2025wan}, a pretrained image-to-video generation framework, consisting of
a variational autoencoder (VAE)~\citep{kingma2013auto}, a transformer-based backbone~\citep{vaswani2017attention} and a text encoder (\textit{i.e.}, umt5-xxl~\citep{chung2023unimax}).
Given a training video $V$, the VAE encoder maps it into a temporally and spatially compressed latent representation $\mathbf{z}_0$, which is used as the clean video latent in diffusion training. 
In Wan2.2, the transformer-based velocity prediction network $\Psi_\theta$ is trained
with Flow Matching (FM)~\citep{lipman2023flow}.
Specifically, given $\mathbf{z}_{0}$ and Gaussian noise $\epsilon \sim \mathcal{N}(\mathbf{0}, \mathbf{I})$, a noisy latent $\mathbf{z}_{\sigma}$
at noise level $\sigma \in [0,1]$ is constructed as $\mathbf{z}_{\sigma} = (1-\sigma)\mathbf{z}_0 + \sigma\epsilon$.
The model is optimized to predict the target velocity with the condition $c$, which is formulated as: 
\begin{align}
\label{eq:loss}
\mathcal{L}_{\mathrm{FM}}
=
\mathbb{E}_{\mathbf{z}_{0},\epsilon,\sigma,c}
\left[
\left\|
\Psi_{\theta}(\mathbf{z}_{\sigma},\sigma,c)
-
(\epsilon-\mathbf{z}_{0})
\right\|_{2}^{2}
\right],
\end{align}
where 
$c$ represents the conditioning features derived from the given text prompt $y$ and the reference image $I_\mathrm{ref}$.

\subsection{Frequency-Aware Identity Encoding}
Wildlife identity is often determined by 
fine-grained local appearance cues, such as fur texture, stripe boundaries, spot configurations, and sharp contour transitions.
However, directly extracting high-frequency responses from the raw reference image is unreliable in natural environments, where background regions such as grass, leaves, shadows, and clutter can also produce strong high-frequency signals. 
To focus the identity presentation on subject-centric appearance, we first isolate the wildlife foreground and then compute its corresponding high-frequency residual map.
Specifically, given a reference image $I_{\mathrm{ref}}$, we apply a foreground extraction operator $f_{\mathrm{seg}}(\cdot)$ on $I_{\mathrm{ref}}$, followed by a spatial high-pass operator $\mathcal{B}(\cdot)$ with a Gaussian smoothing kernel $\mathcal{K}$, to obtain the high-frequency residual map $I_{\mathrm{hf}}$. The formulation is as follows:

\begin{equation}
\label{eq:frequency_extraction}
I_{\mathrm{seg}} = f_{\mathrm{seg}}(I_{\mathrm{ref}}),
\quad
I_{\mathrm{hf}} = \mathcal{B}(I_{\mathrm{seg}})
= I_{\mathrm{seg}} - \mathcal{K} \ast I_{\mathrm{seg}}.
\end{equation}

Using $I_{\mathrm{hf}}$ alone for identity encoding is insufficient, because the residual map discards much of the foreground context and provides limited spatial support for interpreting the extracted high-frequency responses.  
We therefore use $I_{\mathrm{seg}}$ and $I_{\mathrm{hf}}$ as complementary identity cues: $I_{\mathrm{seg}}$ preserves foreground semantics, body structure, and spatial context, while $I_{\mathrm{hf}}$ emphasizes local texture changes and fine residual patterns.
Since $I_{\mathrm{seg}}$ remains close to the natural image distribution while $I_{\mathrm{hf}}$ follows residual image statistics, we adopt a dual-branch identity encoder:
\begin{equation}
\label{eq:identity_encoding}
T_{\mathrm{id}}
=
\mathcal{G}_{\mathrm{id}}(I_{\mathrm{seg}}, I_{\mathrm{hf}})
=
\mathcal{F}\!\left(
\mathcal{E}_{\mathrm{vis}}(I_{\mathrm{seg}})
\oplus
\mathcal{E}_{\mathrm{hf}}(I_{\mathrm{hf}})
\right),
\end{equation}
where $\mathcal{E}_{\mathrm{vis}}$ is a frozen DINO encoder~\citep{simeoni2025dinov3} for foreground visual tokens, $\mathcal{E}_{\mathrm{hf}}$ is a lightweight trainable CNN encoder for high-frequency texture tokens. $\oplus$ denotes feature-wise concatenation along the channel dimension after token alignment, and $\mathcal{F}(\cdot)$ denotes a two-layer MLP-based fusion head with normalization.

\subsection{Identity Conditioning}
After obtaining the identity tokens $T_{\mathrm{id}}$, we inject them into the frozen Wan I2V backbone to strengthen individual identity preservation during generation. 
Our design follows a simple principle: the native Wan model should remain responsible for coarse appearance, structural layout, motion dynamics, and prompt following, while the additional identity branch should provide complementary fine-grained cues that are otherwise difficult to preserve. 
We therefore retain the original image-conditioned generation path of Wan and augment the conditioning space of selected DiT blocks with identity-aware tokens derived from the reference image.

Given a text prompt $y$, the frozen text encoder produces text tokens $T_{\mathrm{txt}}$ that provide motion, scene, and semantic guidance, while the identity tokens $T_{\mathrm{id}}$ summarize fine-grained wildlife-specific appearance cues.
We then project the identity tokens into the Wan text-conditioning feature space and concatenate them with the text tokens, as in Eq.~\eqref{eq:conditioning_sequence}.
Here, $P_{\mathrm{id}}(\cdot)$ is a token-wise two-layer MLP projector, applied to the token feature dimension, and $[\cdot;\cdot]$ denotes token-wise concatenation.
\begin{equation}
\label{eq:conditioning_sequence}
T_{\mathrm{txt}} = \mathrm{TextEncoder}(y),
\quad
T_{\mathrm{cond}} = [\,T_{\mathrm{txt}};\,P_{\mathrm{id}}(T_{\mathrm{id}})\,]
\end{equation}

Here, we inject $T_{\mathrm{cond}}$ only into a selected set $\mathcal{S}$ of middle and late DiT blocks, since early blocks mainly establish coarse layout and temporal structure, whereas middle and late blocks are better suited for refining appearance details and identity-sensitive textures.
For each selected block $l\in\mathcal{S}$, let $X^{(l)}$ denote the incoming visual tokens. 
We update them through the native cross-attention layer:
\begin{equation}
\label{eq:identity_cross_attention}
\tilde{X}^{(l)}
=
X^{(l)}
+
\mathrm{Attn}^{(l)}
\!\left(
X^{(l)}, T_{\mathrm{cond}}
\right),
\quad l\in\mathcal{S},
\end{equation}
where $\mathrm{Attn}^{(l)}(\cdot,\cdot)$ denotes the cross-attention operation of the $l$-th DiT block, with visual tokens as queries and $T_{\mathrm{cond}}$ as keys and values.
For blocks $l\notin\mathcal{S}$, the original Wan conditioning unchanged.

\subsection{Optimization}
WildIcon is 
trained with the FM objective in Eq.~\eqref{eq:loss}.
To further encourage individual-level appearance preservation, we introduce an additional identity loss.
Specifically, given the predicted velocity, we estimate the clean latent as 
$\hat{\mathbf{z}}_{0}=\mathbf{z}_{\sigma}-\sigma\Psi_{\theta}(\mathbf{z}_{\sigma},\sigma,c)$ 
and decode it into $M$ sampled predicted frames $\{\hat I_m\}_{m=1}^{M}$.
Using a frozen ReID encoder $\Phi(\cdot)$, the identity loss is defined as:
\begin{align}
\mathcal{L}_{\mathrm{id}}
=
\frac{1}{M}
\sum_{m=1}^{M}
\left[
1-
\operatorname{sim}
\left(
\Phi(\hat I_m),
\Phi(I_{\mathrm{ref}})
\right)
\right],
\end{align}
where $\operatorname{sim}(u,v)=\frac{u^\top v}{\|u\|\,\|v\|}$ denotes the cosine similarity in the ReID feature space. $\mathcal{L}_{\mathrm{id}}$ encourages the decoded frames to remain close to the reference image in an identity-discriminative feature space. 
The final training objective is:
\begin{align}
\mathcal{L}
=
\mathcal{L}_{\mathrm{FM}}
+
\lambda_{\mathrm{id}}\mathcal{L}_{\mathrm{id}},
\end{align}
where $\lambda_{\mathrm{id}}$ controls the trade-off between video generation quality and identity preservation. 

\section{Experiments}
\label{sec:exp}
In this section, we conduct qualitative and quantitative evaluations of popular I2V generators and our proposed WildIcon to assess their ability to maintain individual-level wildlife consistency.

\subsection{Experimental Settings}

\noindent\textbf{Implementation Details.}
We build our model on Wan2.2-I2V-A14B~\cite{wan2025wan2.2}, which generates 81-frame video clips at $832\times480$ resolution. 
For training, we employ the Adam optimizer with a learning rate of $10^{-4}$. 
We use a per-GPU batch size of $1$ with gradient accumulation over $4$ steps (effective batch size $4$). 
All experiments are conducted on NVIDIA RTX PRO 6000 Blackwell (96GB) GPUs. 

\noindent\textbf{Baselines.}
We compare WildIcon with two groups of baselines.
First, we include representative general I2V and visual-consistency baselines, including Wan2.2 variants (I2V-A14B and TI2V-5B)~\citep{wan2025wan2.2}, CogVideoX1.5-5B~\citep{yang2025cogvideox}, and ConsistI2V~\citep{ren2024consistiv}. 
Second, we compare with representative subject- or identity-consistent video generation methods, including Phantom~\citep{liu2025phantom} and ConsisID~\citep{yuan2025identity}. 

\noindent\textbf{Evaluation metrics.}
An ideal I2V generator should not only align with the text prompt but also preserve the reference subject throughout the generated video. 
Accurately quantifying appearance drift is important for evaluating fine-grained video generation. 
Therefore, we evaluate our method using VBench-based metrics~\cite{huang2024vbench, huang2025vbench++} together with complementary video-generation metrics.
(1) I2V Subject: Cosine similarity between DINO~\cite{simeoni2025dinov3} features of the input image and the generated frames, measuring how well the reference subject is preserved in the generated video.
(2) I2V Background: DreamSim~\cite{fu2023dreamsim} feature similarity between the input image and generated frames, assessing the visual consistency of the scene/background.
(3) Subject Consistency: Average cosine similarity of DINO features across consecutive frames, evaluating subject appearance consistency throughout the video.
(4) Text Relevance: We use CLIPScore~\citep{hessel2021clipscore} to measure the similarity between the generated videos and the input prompts.
(5) Motion Smoothness: Motion prior score derived from a video frame interpolation model~\cite{li2023amt}, assessing whether the generated motion remains smooth.
(6) FVD: Fréchet Video Distance~\cite{unterthiner2018towards} measures the distributional distance between generated and real videos in the feature space of a pretrained video recognition network. 
Lower FVD indicates better overall spatio-temporal realism.

\noindent\textbf{Benchmark.}
To evaluate whether a generator can preserve the identity of the same wildlife individual while following motion and scene instructions, we construct a fixed wildlife generation benchmark from WildlifeVid. 
The benchmark covers 10 species, with 10 identities per species. 
For each identity, we select 1 reference image and pair it with 2 text prompts, resulting in 200 generation cases in total.
All benchmark identities and their associated clips are excluded from WildlifeVid fine-tuning.
All methods are evaluated on this fixed benchmark under identical output resolution and video duration.

\begin{table*}[t]
\centering
\fontsize{9pt}{9pt}\selectfont
\caption{Quantitative comparison on wildlife individual-consistent image-to-video generation. We compare WildIcon with representative I2V generation models.}
\label{tab:results_quantitative}
\begin{tabular}{@{}lcccccc@{}}
\toprule
Method & \makecell{I2V\\Subject $\uparrow$} & \makecell{I2V\\Background $\uparrow$} & \makecell{Subject\\Consistency $\uparrow$} & \makecell{Motion\\Smoothness $\uparrow$} & \makecell{Text \\ Relevance $\uparrow$} & FVD $\downarrow$ \\ \midrule
CogVideoX-5B~\citep{yang2025cogvideox} & 82.18 & 86.74 & 89.92 & 95.62 & 25.63 & 184 \\
Wan2.2-5B~\citep{wan2025wan2.2} & 86.02 & 88.01 & 90.16 & 95.94 & 26.57 & 175 \\
Wan2.2-A14B~\citep{wan2025wan2.2} & 86.23 & 88.27 & 90.79 & 96.43 & \textbf{26.58} & 152 \\
ConsistI2V~\citep{ren2024consistiv} & 83.71 & 86.18 & 88.58 & 95.08 & 25.21 & 188 \\
ConsisID~\citep{yuan2025identity} & 84.06 & 86.29 & 89.72 & 95.56 & 25.98 & 168 \\
Phantom~\citep{liu2025phantom} & 85.64 & 87.53 & 91.02 & 96.11 & 26.25 & 172 \\ \midrule
WildIcon (Ours) & \textbf{87.40} & \textbf{89.36} & \textbf{92.94} & \textbf{96.85} & 26.38 & \textbf{139} \\ \bottomrule
\end{tabular}
\end{table*}

\begin{figure*}[ht]
    \centering
    \includegraphics[width=0.90\columnwidth]{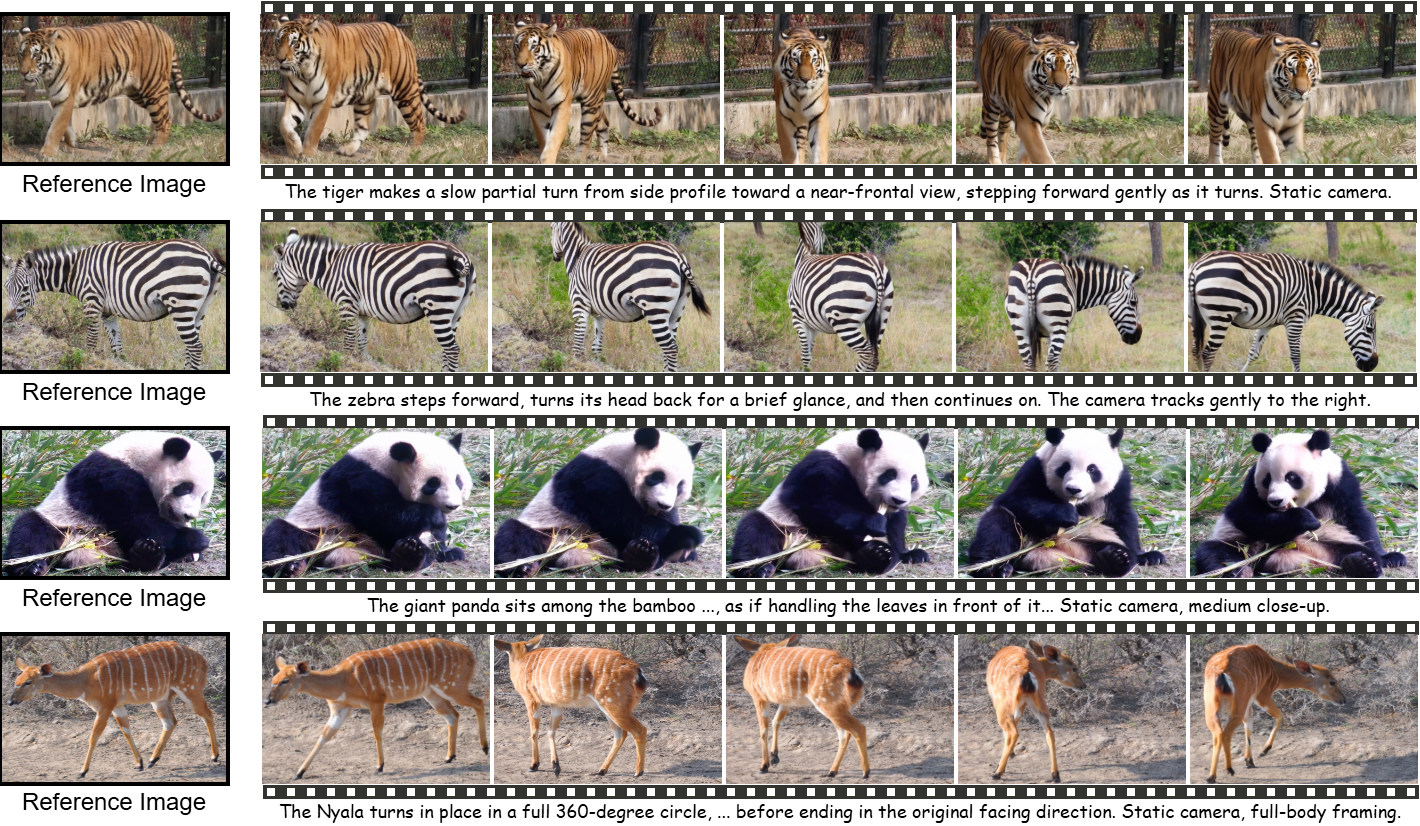}
    \caption{\textbf{Examples of individual-consistent video generation by our WildIcon.}
    Given a reference image and a motion prompt, WildIcon generates videos that follow the specified motion while maintaining fine-grained appearance cues visible in the reference image across frames.} 
    \label{fig:qualitative_examples}
\end{figure*}

\subsection{Experimental Results}
\label{subsec:results}
\noindent\textbf{Quantitative Results.}
As shown in Table~\ref{tab:results_quantitative}, WildIcon achieves the best overall performance on our fixed wildlife generation benchmark. 
Compared with representative general-purpose I2V baselines and prior subject- or identity-consistent generation methods, our model delivers the clearest gains on identity preservation, improving Subject Consistency from 91.02\% to 92.94\%.
These results suggest that WildIcon better preserves identity-critical wildlife appearance cues throughout the generated video while maintaining stable scene coherence and smooth temporal dynamics. 
Notably, these gains are achieved without sacrificing text relevance or motion quality, showing that stronger reference-conditioned appearance consistency remains compatible with the general generation capability of the pretrained I2V backbone.

\noindent\textbf{Qualitative Results.}
Figure~\ref{fig:qualitative_examples} presents representative examples of wildlife individual-consistent video generation. 
The generated videos exhibit consistent fine-grained identity cues across frames, such as fur texture, stripe boundaries, and local contour details, while maintaining realistic motion and scene coherence.

\begin{figure}[ht]
    \centering
    \includegraphics[width=0.90\linewidth]{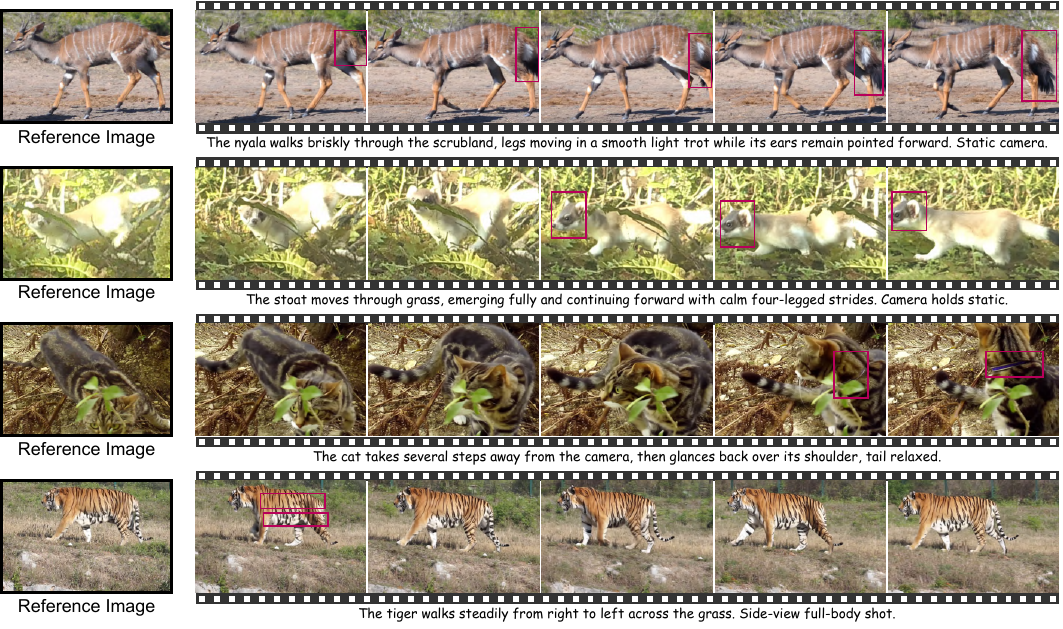}
    \caption{\textbf{Representative challenging cases under incomplete or ambiguous reference evidence.}
    Red boxes highlight local regions where overlap, occlusion, or scene structures in the reference image lead to unsupported or drifting appearance details during generation.}
    \label{fig:failure_cases}
\end{figure}

\noindent\textbf{Challenging cases.}
Figure~\ref{fig:failure_cases} further illustrates several challenging cases under incomplete or ambiguous reference evidence.
For the nyala, its tail overlaps with another individual's head in the reference image, causing the overlap to be partially carried into the generated sequence.
For the stoat and cat, vegetation obscures parts of the face; when the generated motion reveals regions that are not clearly visible in the reference, local appearance may drift, and unsupported details can emerge, such as the collar-like artifact in the cat example.
For the tiger, thin fence wires crossing the animal are initially retained in the foreground, likely because these high-frequency structures are spatially entangled with the animal in the reference image.
As the animal moves away from their original image locations, these artifacts gradually weaken.
Overall, these examples show that generation is most reliable when the reference provides clear identity evidence and the requested motion does not require substantial extrapolation of unseen identity-bearing regions.

\begin{table*}[t]
\centering
\fontsize{9pt}{9pt}\selectfont
\setlength{\tabcolsep}{1.0pt}
\renewcommand{\arraystretch}{1.08}
\caption{\textbf{Ablation studies of WildIcon.}
Subj., Back., Cons., and Mot. denote I2V Subject, I2V Background, Subject Consistency, and Motion Smoothness.
FG denotes foreground-only input, HF denotes high-frequency-only input, LoRA-FT denotes LoRA fine-tuning of Wan2.2 on WildlifeVid, and Full denotes the default WildIcon setting.
Default configurations are marked in \colorbox{baselinecolor}{gray}.
}
\label{tab:results_ablation}
\begin{minipage}[t]{0.315\textwidth}
\centering
(a) Identity branch input.
\vspace{0.35em}

\begin{tabular}{@{}lcccc@{}}
\toprule
Setting & Subj. & Back. & Cons. & Mot. \\
\midrule
FG-only & 86.81 & 88.39 & 91.33 & 96.53 \\
HF-only & 86.69 & 88.34 & 91.48 & 96.28 \\
\rowcolor{baselinecolor}
FG+HF & \textbf{87.40} & \textbf{89.36} & \textbf{92.94} & \textbf{96.85} \\
\bottomrule
\end{tabular}
\end{minipage}
\hfill
\begin{minipage}[t]{0.315\textwidth}
\centering
(b) Identity injection strategy.
\vspace{0.35em}

\begin{tabular}{@{}lcccc@{}}
\toprule
Setting & Subj. & Back. & Cons. & Mot. \\
\midrule
Early & 86.74 & 88.86 & 91.06 & 96.24 \\
All & 86.82 & 88.74 & 91.29 & 96.04 \\
\rowcolor{baselinecolor}
Mid\&late & \textbf{87.40} & \textbf{89.36} & \textbf{92.94} & \textbf{96.85} \\
\bottomrule
\end{tabular}
\end{minipage}
\hfill
\begin{minipage}[t]{0.315\textwidth}
\centering
(c) Training/data effect.
\vspace{0.35em}

\begin{tabular}{@{}lcccc@{}}
\toprule
Setting & Subj. & Back. & Cons. & Mot. \\
\midrule
Frozen & 86.23 & 88.27 & 90.79 & 96.43 \\
LoRA-FT & 86.95 & 88.69 & 91.63 & 96.45 \\
\rowcolor{baselinecolor}
Full & \textbf{87.40} & \textbf{89.36} & \textbf{92.94} & \textbf{96.85} \\
\bottomrule
\end{tabular}
\end{minipage}
\end{table*}

\subsection{Ablation studies}
\noindent \textbf{Effect of Key Components.} 
Table~\ref{tab:results_ablation}(a) studies the input design of the identity branch.
Results show that combining foreground and high-frequency inputs outperforms either cue alone, confirming that foreground semantic context and high-frequency residual details provide complementary identity cues.
Table~\ref{tab:results_ablation}(b) evaluates where identity conditioning should be injected.
The default middle-to-late injection strategy outperforms early-block and all-block conditioning, suggesting that identity cues are most effective when introduced during feature refinement, rather than during early layout formation or throughout all blocks.

\noindent \textbf{Effect of WildlifeVid Training.} 
Table~\ref{tab:results_ablation}(c) examines the effect of training on WildlifeVid.
LoRA fine-tuning of Wan2.2 improves over the frozen baseline, indicating that WildlifeVid provides useful supervision for wildlife I2V generation.
However, the full WildIcon setting achieves stronger results than LoRA fine-tuning alone.
This underscores that our architectural design is the key factor behind the performance boost.

\section{WildIcon for Animal ReID}
\label{sec:animal_reid}
\begin{table*}[t]
\centering
\fontsize{9pt}{9pt}\selectfont
\setlength{\tabcolsep}{3.8pt}
\caption{\textbf{Downstream animal ReID results on five species.}
All settings use the same real training split, synthetic-data construction pipeline, and unchanged real-only query/gallery splits.
We report mAP and Rank-1 (R1) (\%) on each species. Best results within each backbone block are shown in bold.
Results are averaged over ten runs; 95\% confidence intervals are reported in the Appendix.}
\label{tab:animal_reid_main}
\begin{tabular}{@{}llcccccccccc@{}}
\toprule
\multirow{2}{*}{Model} & \multirow{2}{*}{Setting}
& \multicolumn{2}{c}{Tiger}
& \multicolumn{2}{c}{Nyala}
& \multicolumn{2}{c}{Cow}
& \multicolumn{2}{c}{Panda}
& \multicolumn{2}{c}{Stoat} \\
\cmidrule(lr){3-4}
\cmidrule(lr){5-6}
\cmidrule(lr){7-8}
\cmidrule(lr){9-10}
\cmidrule(lr){11-12}
& & mAP & R1 & mAP & R1 & mAP & R1 & mAP & R1 & mAP & R1 \\
\midrule
\multirow{4}{*}{SigLIP-FT~\cite{zhai2023sigmoid}}
& Real only            & 55.13 & 90.21 & 25.84 & 41.83 & 62.57 & 91.53 & 28.97 & 45.43 & 61.16 & 92.06 \\
& + Data Aug.       & 55.41 & 90.08 & 26.11 & 42.06 & 62.74 & 91.47 & 29.22 & 45.31 & 61.38 & 92.14 \\
& + Wan synth.      & 57.24 & 91.36 & 29.08 & 43.02 & 64.08 & 92.09 & 31.14 & 47.82 & 62.94 & 92.57 \\
& + Ours synth.     & \textbf{59.72} & \textbf{93.41}
                    & \textbf{32.01} & \textbf{46.36}
                    & \textbf{65.84} & \textbf{92.11}
                    & \textbf{32.26} & \textbf{49.36}
                    & \textbf{65.18} & \textbf{94.71} \\
\midrule
\multirow{4}{*}{DINO-FT~\cite{simeoni2025dinov3}}
& Real only            & 54.72 & 92.61 & 22.94 & 35.87 & 66.31 & 95.08 & 29.34 & 44.56 & 63.07 & 93.02 \\
& + Data Aug.       & 54.89 & 92.40 & 23.06 & 36.02 & 66.47 & 95.03 & 29.58 & 44.61 & 63.01 & 92.94 \\
& + Wan synth.      & 56.38 & 93.44 & 25.41 & 38.09 & 67.73 & 95.96 & 30.86 & 46.18 & 64.47 & 94.01 \\
& + Ours synth.     & \textbf{58.21} & \textbf{94.27}
                    & \textbf{27.86} & \textbf{40.72}
                    & \textbf{69.12} & \textbf{96.54}
                    & \textbf{31.94} & \textbf{47.63}
                    & \textbf{66.13} & \textbf{94.87 }\\ \midrule
\multirow{4}{*}{CLIP-ReID~\cite{li2023clip}}
& Real only            & 54.10 & 94.22 & 18.49 & 28.96 & 68.43 & 97.14 & 29.71 & 42.10 & 62.83 & 91.59 \\
& + Data Aug.       & 54.02 & 94.10 & 18.88 & 29.11 & 68.31 & 97.06 & 29.95 & 42.28 & 63.01 & 91.54 \\
& + Wan synth.      & 56.41 & 95.03 & 21.72 & 31.64 & 69.57 & 97.61 & 31.38 & 44.08 & 64.11 & 92.42 \\
& + Ours synth.     & \textbf{58.36} & \textbf{95.96}
                    & \textbf{24.58} & \textbf{34.29}
                    & \textbf{71.04} & \textbf{97.85}
                    & \textbf{32.47} & \textbf{45.96}
                    & \textbf{65.72} & \textbf{93.36} \\
\bottomrule
\end{tabular}
\end{table*}
To further validate WildIcon, we evaluate on animal ReID, a fine-grained retrieval task for matching the same individual across different images.
This experiment tests whether videos synthesized by different generators can provide useful training samples for downstream individual-level recognition. 
We use mean Average Precision (mAP)~\cite{wu2017rgbinfrared} and Cumulative Matching Characteristics (CMC)~\cite{moon2001computational} as evaluation metrics. 
mAP measures overall retrieval performance by averaging precision across all query samples in the query set, while CMC measures the proportion of queries for which the correct identity appears within the top-$k$ retrieved results.
We report mAP and CMC at Rank-1 ($\%$), averaged over 10 runs, and provide the corresponding 95\% confidence intervals in the Appendix.

Our evaluation follows a fixed synthetic-data construction pipeline. 
Starting from a curated real training subset, we use different I2V generators to synthesize videos under the same reference images and prompts, and then sample frames from the generated videos as synthetic training candidates.
Each candidate is compared with its source reference image using a fixed DINO-based similarity filter, and only high-similarity samples are retained.
We further apply a per-identity balanced selection rule to prevent synthetic samples from amplifying identity imbalance.
This design supports two controlled comparisons. 
(1) To examine whether performance gains arise merely from increased data volume or from the additional supervision provided by reference-consistent generated samples, we include a classical data augmentation baseline with the same additional-sample budget.
(2) To assess whether the generated data provides generally useful supervision, we evaluate all training sets with multiple ReID backbones, including SigLIP-FT~\citep{zhai2023sigmoid}, DINO-FT~\citep{simeoni2025dinov3}, and CLIP-ReID~\cite{li2023clip}. 
All remaining factors are held constant across generators, including the reference-image subset, prompt set, frame-extraction schedule, similarity-filtering rule, synthetic-data budget, ReID training setup, and the unchanged real-only query/gallery split.

As shown in Table~\ref{tab:animal_reid_main}, adding synthetic data consistently improves ReID performance over the real-only setting across all backbones.
In contrast, classical data augmentation provides only marginal and less stable gains, indicating that the improvement is not simply caused by increasing the number of training images.
Under the same synthetic-data budget, samples generated by WildIcon outperform those generated by Wan2.2 across the evaluated ReID models, indicating that downstream improvements are primarily driven by the quality of identity preservation in generated samples rather than by the presence of synthetic data itself.
Moreover, consistent improvements across different ReID architectures show that WildIcon provides reusable supervisory signals, supporting individual-consistent I2V generation as a practical data enrichment strategy for animal ReID.
A generated-to-real retrieval evaluation further shows that WildIcon outperforms both Wan2.2 and WildlifeVid LoRA fine-tuning across all five datasets, while replacing DINOv3 with SigLIP for synthetic-sample selection preserves the downstream advantage (Appendix~\ref{app:additional_reid_evals}).

\section{Conclusion}
\label{sec:conclusion}
In this paper, we explored individual-consistent I2V generation for wildlife as a practical direction for mitigating the data scarcity that limits wildlife identification.  
To support this task, we proposed WildlifeVid, a wildlife-centric video dataset with individual-level identity labels for training and evaluating individual-consistent I2V models. 
We further proposed WildIcon, a frequency-aware I2V framework that uses foreground-filtered high-frequency cues to better maintain identity-relevant appearance supported by the reference image.
Experiments showed that WildIcon achieves stronger identity consistency than existing baselines, and that its generated samples consistently improve downstream animal ReID across five species. 
These results suggest that individual-consistent wildlife I2V generation can serve as a practical tool for enriching training data for wildlife identification and supporting broader biodiversity monitoring efforts when real observations are limited.

\section*{Acknowledgements}
This research was supported by the Ministry of Business, Innovation and Employment (MBIE) Smart Ideas project ``Stoat Re-Identification Powered by Artificial Intelligence for Effective Pest Management in Aotearoa'' (UOA2505).

\bibliographystyle{unsrtnat}
\bibliography{references}

\newpage
\appendix

\section{Implementation and Evaluation Details}
\label{app:exp}
\subsection{Reproducibility}
We open-source the implementation of WildIcon at \url{https://github.com/ML-4-SocialGood/WildIcon}.
The repository contains the code needed to reproduce our individual-consistent wildlife I2V framework, including the frequency-aware identity encoder, foreground-isolated high-frequency extraction, identity-token projection, selective identity conditioning, training and inference pipelines, and the evaluation scripts used in our experiments.

\noindent\textbf{Code.}
The released code is organized around the main components of WildIcon. 
It includes: 
(1)~a foreground extraction and high-frequency preprocessing module that constructs $I_{\mathrm{seg}}$ and $I_{\mathrm{hf}}$ from a reference image; 
(2)~a dual-branch identity encoder that encodes foreground appearance and high-frequency residual cues into identity tokens; 
(3)~an identity-conditioning module that projects the identity tokens into the conditioning space and injects them into selected middle-to-late DiT blocks; 
and (4)~training and inference scripts for generating wildlife videos from a reference image and a standardized text prompt. 
The repository also provides environment requirements, example commands, and configuration files for reproducing the main experimental settings.

\noindent\textbf{Dataset and benchmark curation.}
To support reproducibility of WildlifeVid and the evaluation benchmark, the repository includes the metadata construction and curation utilities used in our pipeline. 
These scripts cover source-level metadata processing, clip-level filtering, standardized prompt organization, and identity-merging utilities. 

\noindent\textbf{Evaluation.}
The repository contains the evaluation scripts used to compute the video-generation metrics reported in the paper, including subject preservation, background consistency, subject consistency, motion smoothness, text relevance, and FVD evaluation. 

\subsection{Data Accessibility and Release}
\label{appendix:accessibility}
WildlifeVid is constructed by curating and re-annotating clips from existing wildlife video resources, including videos hosted on public platforms where applicable. 
We acknowledge that such source videos may be subject to copyright protection, platform terms of service, and the usage conditions of the original datasets.
To respect these terms and mitigate copyright, redistribution, and ecological-sensitivity concerns, we do not redistribute the original videos or raw video frames. 
Instead, we publicly release only the annotations, which do not contain any raw video content. 
All derived data is non-identifiable and used solely for research purposes.
The WildlifeVid full annotation files are distributed separately as supplementary material.
Our newly created WildlifeVid annotations, prompts, scripts, and statistics are released under CC BY 4.0.

\subsection{Animal ReID Datasets}
\label{app:animal-reid-datasets}
We evaluate the downstream utility of WildIcon on five animal ReID datasets covering four public datasets, Tiger~\citep{li2020atrw}, Nyala~\citep{dlamini2020automated}, Cow~\citep{fu2022lightweight}, Panda~\citep{wang2021giant}, and our proprietary camera-trap Stoat dataset. 
For each dataset, we organize the data into a processed ReID split consisting of training, query, and gallery sets. 
The resulting dataset statistics are summarized in Table~\ref{tab:animal_reid_datasets}. 

\begin{table*}[ht]
\centering
\renewcommand{\arraystretch}{1.08}
\caption{\textbf{Animal ReID datasets used for downstream evaluation.}
We report the processed ReID splits used in our experiments. 
These statistics correspond to our curated splits and may differ from the raw dataset statistics reported in the original papers.}
\label{tab:animal_reid_datasets}
\begin{tabular}{@{}llrrrrrrrr@{}}
\toprule
\multirow{2}{*}{Animal} 
& \multirow{2}{*}{Dataset}
& \multicolumn{2}{c}{Train}
& \multicolumn{2}{c}{Query}
& \multicolumn{2}{c}{Gallery}
& \multicolumn{2}{c}{Total} \\
\cmidrule(lr){3-4}
\cmidrule(lr){5-6}
\cmidrule(lr){7-8}
\cmidrule(l){9-10}
& 
& Imgs & IDs 
& Imgs & IDs 
& Imgs & IDs 
& Imgs & IDs \\
\midrule
Tiger & ATRW~\citep{li2020atrw}
& 3,678 & 148 
& 424 & 33 
& 521 & 33 
& 4,623 & 181 \\

Nyala & Nyala~\citep{dlamini2020automated} 
& 1,213 & 179 
& 354 & 58 
& 375 & 58 
& 1,942 & 237 \\

Cow & CowDataset~\citep{fu2022lightweight} 
& 919 & 8 
& 210 & 5 
& 356 & 5 
& 1,485 & 13 \\

Panda & IPanda50~\citep{wang2021giant} 
& 4,114 & 31 
& 1,022 & 19 
& 1,738 & 19 
& 6,874 & 50 \\

Stoat & Stoat (Proprietary) 
& 1,390 & 37 
& 403 & 46 
& 780 & 46 
& 2,573 & 83 \\
\bottomrule
\end{tabular}
\end{table*}

\noindent\textbf{Animal-centric preprocessing.}
All real training, query, gallery, and synthetic-filtering inputs are processed as animal-centric crops using the same detection and preprocessing pipeline.
The similarity used for synthetic-sample filtering is computed between the animal crop of each generated frame and the corresponding animal crop of its real reference image, rather than between full-scene images.
This reduces the influence of background and scene context in both synthetic-sample selection and downstream ReID evaluation.

\subsection{ReID Evaluation with Synthetic Data}
\label{app:reid-synthetic}
To evaluate the downstream utility of generated videos, we construct synthetic training images from videos generated by each I2V model under a fixed pipeline. 
For fairness, all compared methods use the same reference-image pool, text prompts, video-generation settings, frame-sampling rule, similarity-filtering threshold, and final synthetic-data budget. 
The query and gallery sets are kept fixed across all experiments and contain only real images; synthetic samples are used only to augment the training set.

\noindent\textbf{Reference image selection.}
To prevent test-set leakage, reference images are selected from only the real training split of each ReID dataset.
Let $\mathcal{D}^{\mathrm{train}}_i$ denote the real training images of identity $i$.
We construct a reference pool $\mathcal{R}_i \subseteq \mathcal{D}^{\mathrm{train}}_i$ by excluding images that are unsuitable as I2V references, such as severely blurred, low-visibility nighttime, or infrared images with weak visible identity cues.
This filtering is applied only to the reference pool for video generation; the original real, query, and gallery splits remain unchanged.
The same reference pool is used for all compared generation methods.

\noindent\textbf{Video generation protocol.}
For each selected reference image $r \in \mathcal{R}_i$, we generate two videos. 
The prompt is designed to encourage mild motion and viewpoint variation starting from the reference pose, such as slight turning, small body movement, or modest camera-view changes, while preserving the identity of the target animal as much as possible. 
All generated videos are processed using the default generation setting of 81 frames at 16 FPS.

\noindent\textbf{Reference-evidence admission guardrail.}
For downstream ReID augmentation, we restrict generation to motion and viewpoint changes that remain supported by the visual evidence in the reference image.
Prompts requiring a full $360^\circ$ rotation, explicit exposure of the opposite side, or a camera orbit around the animal are excluded because they may reveal identity-bearing regions that are not observed in the reference.
Generated candidates are further checked for whether sufficient reference-visible appearance remains comparable to the source image.
Candidates that expose unsupported identity-bearing regions or otherwise become visually uncertain are discarded before the similarity-filtering stage and are not assigned the source identity label.

\noindent\textbf{Frame sampling.}
To convert generated videos into synthetic training images, we sample frames at fixed temporal intervals. 
We discard the first 5 frames and the last 5 frames of each generated video to avoid frames that are overly close to the static input image and frames that may contain end-of-video artifacts, yielding 8 candidate synthetic frames per video.
For an 81-frame video indexed from $0$ to $80$, we use
\[
\mathcal{F}_{\mathrm{samp}}=\{5,15,25,35,45,55,65,75\},
\]

\noindent\textbf{DINO similarity filtering.}
After frame sampling, we filter synthetic candidates based on their appearance consistency with the corresponding source reference image.
Let $\Phi(\cdot)$ denote the frozen DINOv3 image encoder.
For a sampled frame $x$ generated from the reference image $r$, we compute
\[
\bar{\Phi}(x)=\frac{\Phi(x)}{\|\Phi(x)\|_2},
\quad
s(x,r)=\bar{\Phi}(x)^\top \bar{\Phi}(r).
\]
A sampled frame is retained only if $s(x,r)\geq\tau$, where $\tau=0.90$ is fixed for all datasets and generation methods.
This filtering step removes candidates with large appearance deviations from the source reference and retains samples that remain visually close to the reference under the fixed DINO representation.

\noindent\textbf{Synthetic budget control.}
After similarity filtering, directly adding all retained frames could make the synthetic data overly dominant and amplify identity imbalance, because identities with more valid generated frames would receive disproportionately more synthetic samples.
We therefore apply a second-stage budget control procedure to construct a balanced augmented training set.

For each dataset, we first set a target synthetic fraction $\rho$ for the final training set. 
For an identity $i$ with $n_i^{\text{real}}$ real training images, the per-identity synthetic budget is computed as
\[
b_i = \operatorname{round}\!\left(\frac{n_i^{\text{real}} \cdot \rho}{1-\rho}\right),
\]
so that the selected synthetic samples form approximately a fraction $\rho$ of the final real-plus-synthetic training set. 
In our experiments, we use $\rho = 0.2$, corresponding to a target synthetic fraction of approximately $20\%$. 
We also limit each source reference image to a maximum of 2 selected synthetic frames to prevent a single reference image from dominating the augmented set. 
Within each identity, candidate synthetic frames are ranked by DINO similarity to the corresponding source reference image, and the highest-scoring frames are selected subject to the per-identity budget and per-source cap. 
If an identity has fewer valid candidates than its allocated budget, all valid candidates are retained, and the remaining budget is not filled with lower-similarity samples.
Let $\mathcal{A}_i$ denote the selected synthetic samples for identity $i$, and let
\[
\mathcal{A}_{\mathrm{syn}}=\bigcup_i \mathcal{A}_i
\]
denote the final selected synthetic subset across all identities.
The augmented training set is then
\[
\mathcal{D}^{\mathrm{train}}_{\mathrm{aug}}
=
\mathcal{D}^{\mathrm{train}}_{\mathrm{real}}
\cup
\mathcal{A}_{\mathrm{syn}} .
\]
The query and gallery sets remain unchanged throughout all ReID experiments.
They contain only real images and never include synthetic samples.
Therefore, synthetic data are used only to augment the real training set, and all reported ReID results are evaluated on the same fixed real-only query/gallery splits.

\begin{figure*}[t]
    \centering
    \includegraphics[width=1.0\columnwidth]{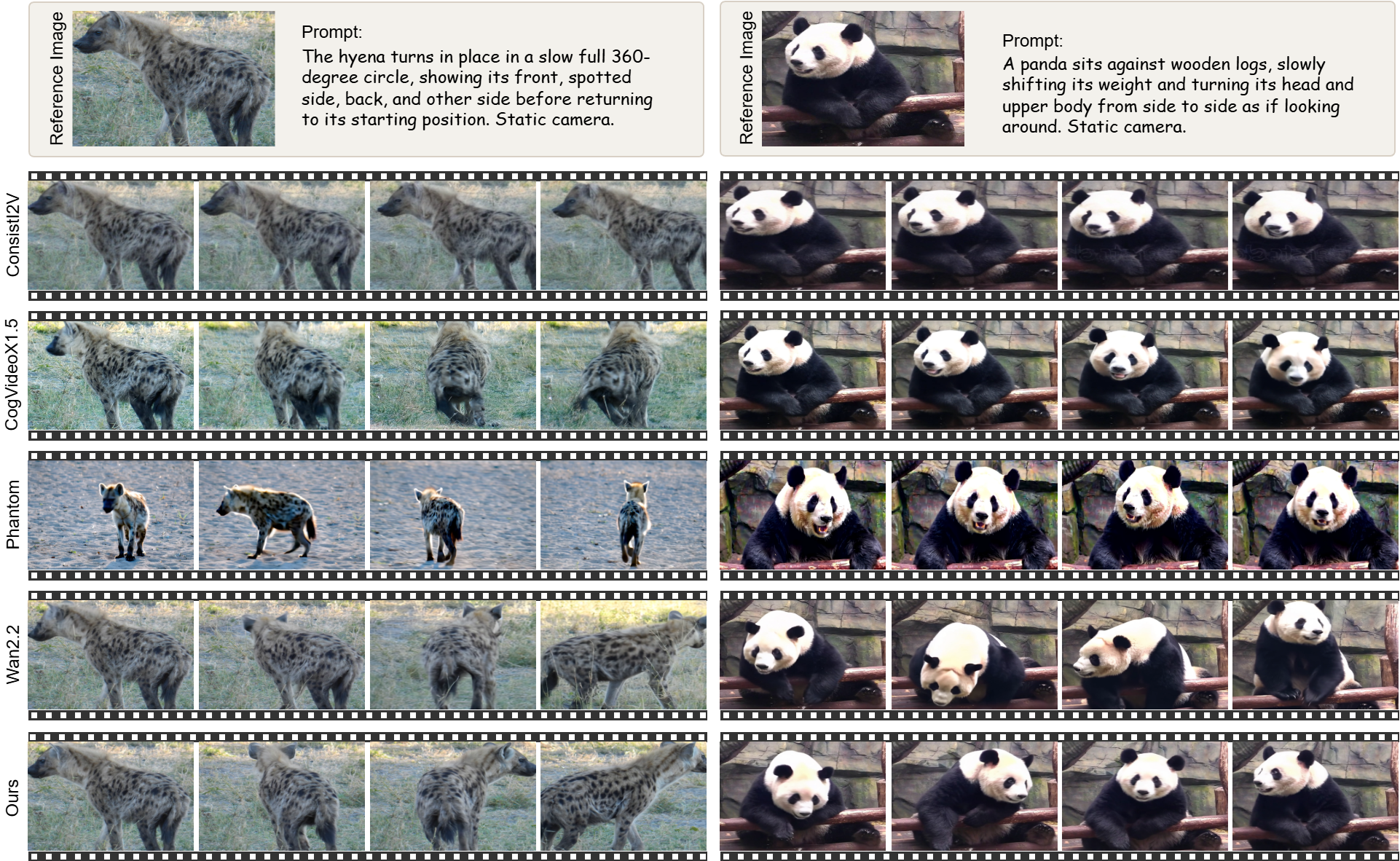}
    \caption{\textbf{Qualitative comparison with representative video generators.}
    Given the same reference image and text prompt, we compare WildIcon with representative general I2V and identity-preserving video generation baselines.} 
    \label{fig:app_qualitative_examples}
\end{figure*}

\begin{table*}[t]
\centering
\fontsize{9pt}{9pt}\selectfont
\caption{\textbf{Quantitative ablation studies of WildIcon.}
We study three factors: the input to the identity branch, the strategy for injecting identity conditioning, and the effect of the WildlifeVid dataset. The final row reports the full WildIcon model as the common reference for all ablations.}
\label{tab:results_ablation_appendix}
\begin{tabular}{@{}lcccccc@{}}
\toprule
Method
& \makecell{I2V\\Subject $\uparrow$}
& \makecell{I2V\\Background $\uparrow$}
& \makecell{Subject\\Consistency $\uparrow$}
& \makecell{Motion\\Smoothness $\uparrow$}
& \makecell{Text\\Relevance $\uparrow$}
& FVD $\downarrow$ \\
\midrule
Wan2.2-A14B               & 86.23 & 88.27 & 90.79 & 96.43 & \textbf{26.58} & 152 \\
\midrule
\multicolumn{7}{@{}l}{\textit{Identity branch input}} \\
Full image                & 86.57 & 88.41 & 91.07 & 96.45 & 26.56 & 159 \\
Foreground only           & 86.81 & 88.39 & 91.33 & 96.53 & 26.49 & 148 \\
High-frequency only       & 86.69 & 88.34 & 91.48 & 96.28 & 26.43 & 150 \\
\midrule
\multicolumn{7}{@{}l}{\textit{Injection strategy}} \\
Early blocks              & 86.74 & 88.86 & 91.06 & 96.24 & 26.47 & 147 \\
All blocks                & 86.82 & 88.74 & 91.29 & 96.04 & 26.24 & 144 \\
\midrule
\multicolumn{7}{@{}l}{\textit{Dataset effectiveness}} \\
Wan2.2-A14B-FT            & 86.95 & 88.69 & 91.63 & 96.45 & 26.55 & 146 \\
\midrule
WildIcon (Ours) & \textbf{87.40} & \textbf{89.36} & \textbf{92.94} & \textbf{96.85} & 26.38 & \textbf{139} \\
\bottomrule
\end{tabular}
\end{table*}

\begin{figure*}[ht]
    \centering
    \includegraphics[width=0.99\columnwidth]{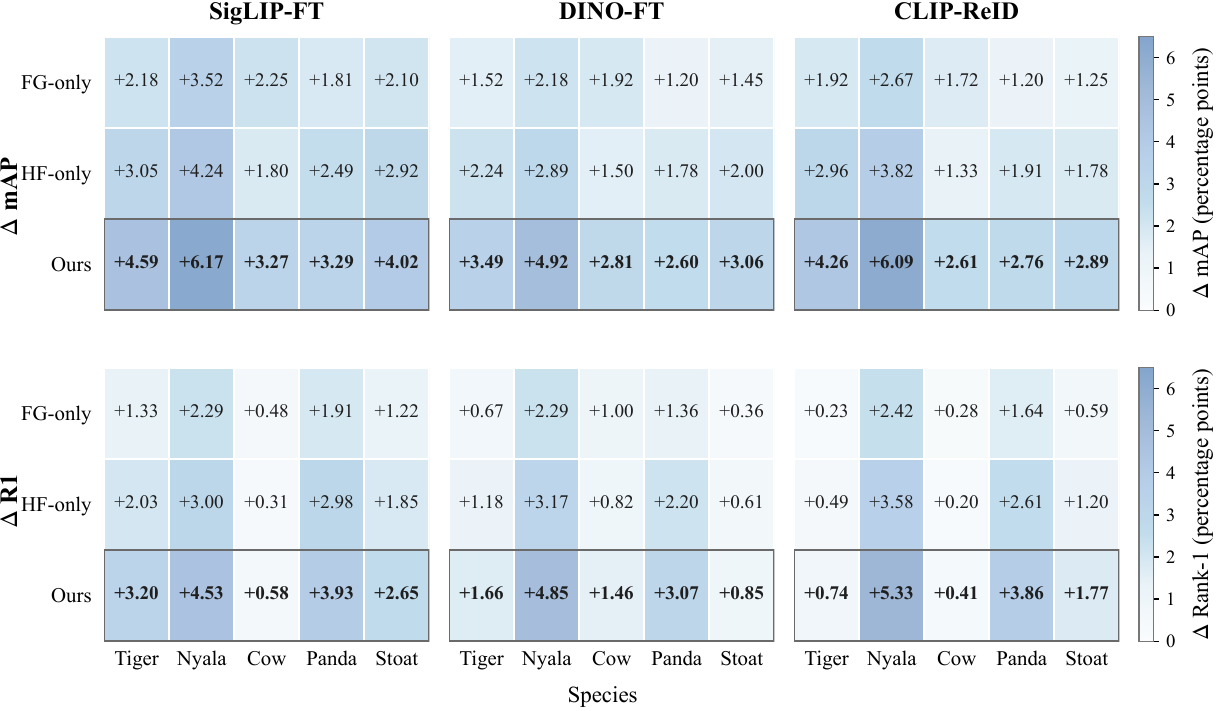}
    \caption{\textbf{Downstream ReID ablation of WildIcon identity-branch inputs.}
    We visualize the absolute performance gains over the real-only baseline for different identity-branch variants, including foreground-only conditioning (FG-only), high-frequency-only conditioning (HF-only), and the full WildIcon identity branch (Ours).
    Darker colors indicate larger downstream ReID gains.}
    \label{fig:appendix_reid_ablation_identity_branch}
\end{figure*}

\section{Additional Results}
\label{app:results}

\subsection{Qualitative comparison with video generation baselines}
Figure~\ref{fig:app_qualitative_examples} complements the quantitative comparison in Table~\ref{tab:results_quantitative}.
Under the same reference images and prompts, existing video generators often produce plausible motion but exhibit identity-cue drift, including blurred stripe or spot boundaries, shifted fur texture, altered local markings, and inconsistent body contours.
These failures indicate that generic image-conditioned or identity-preserving video generators do not explicitly preserve the fine-grained local cues required for individual consistency in wildlife.
WildIcon better maintains these identity-critical patterns across frames while preserving coherent motion and prompt alignment.

\subsection{Additional Ablation Results}
\label{app:ablation_results}
\noindent\textbf{Quantitative ablation studies of WildIcon.}
Table~\ref{tab:results_ablation_appendix} provides a more complete ablation analysis of WildIcon, covering three factors: the identity-branch input, the identity-conditioning injection strategy, and the effect of WildlifeVid training.

First, the identity-branch input ablation shows that directly using the full reference image provides only limited gains over Wan2.2-A14B. 
This suggests that raw reference-image conditioning may introduce background clutter and non-identity visual details that are not useful for preserving wildlife identity. 
In contrast, foreground-only conditioning improves both identity preservation and video realism by suppressing background interference.
High-frequency-only conditioning yields slightly stronger subject consistency than foreground-only conditioning, indicating that local residual cues are useful for maintaining identity-sensitive texture and boundary details. 
However, it performs worse than the full WildIcon setting, because high-frequency residuals alone discard semantic foreground context and spatial support. 
The full WildIcon identity branch achieves the best overall results by combining foreground appearance with foreground-filtered high-frequency cues, confirming that these two sources provide complementary identity evidence.

Second, the injection-strategy ablation shows that injecting identity conditioning into early blocks or all blocks is less effective than the default middle-to-late strategy. 
Early-block injection improves background consistency but provides weaker subject consistency, suggesting that introducing identity cues too early may interfere with coarse layout and motion formation. 
All-block injection improves FVD compared with early-only injection, but it reduces motion smoothness and text relevance, indicating that overly persistent identity conditioning can constrain prompt-driven generation. 
The middle-to-late injection strategy gives the best balance, as these blocks are better suited for refining appearance details and identity-sensitive textures after global structure and motion have been established.

Third, the dataset-effectiveness ablation shows that LoRA fine-tuning Wan2.2-A14B on WildlifeVid improves over the frozen Wan2.2-A14B baseline, showing that WildlifeVid provides useful supervision for wildlife I2V generation. 
However, fine-tuning alone remains below the full WildIcon model across the main identity and realism metrics. 
This indicates that the gains are not solely due to domain-specific fine-tuning data, but also depend on the frequency-aware identity branch and selective identity conditioning.

\begin{figure*}[ht]
    \centering
    \includegraphics[width=0.99\columnwidth]{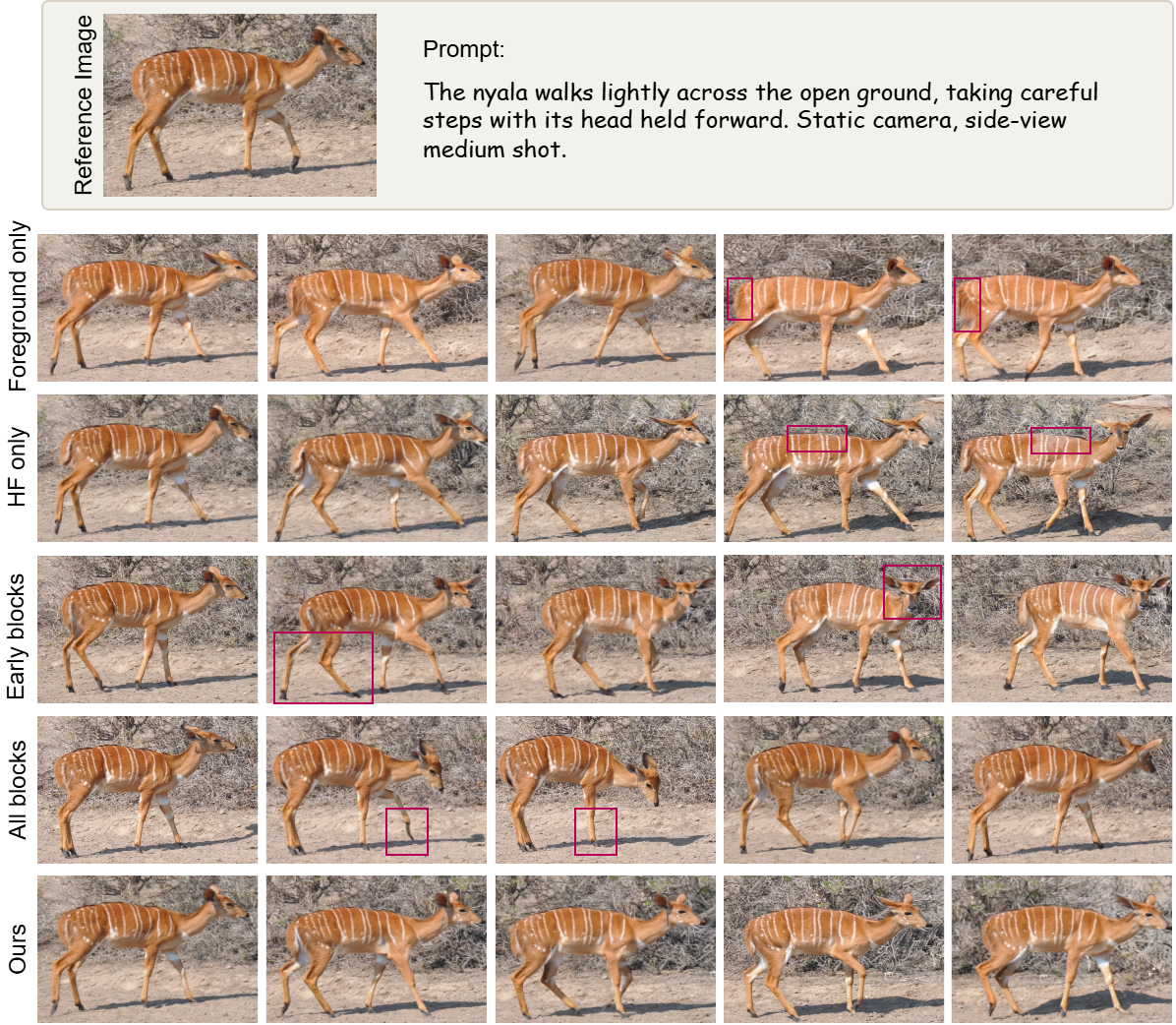}
    \caption{\textbf{Qualitative ablation studies of WildIcon.}
    We compare WildIcon variants under the same reference image and prompt. Red boxes mark representative artifacts, including tail deformation, back texture drift, head appearance changes, motion inconsistency, and limb distortion. The full WildIcon model better preserves local identity cues and structural consistency across frames.} 
    \label{fig:app_abltion_qualitative_examples}
\end{figure*}

\noindent\textbf{Qualitative ablation studies of WildIcon.}
Figure~\ref{fig:app_abltion_qualitative_examples} provides qualitative evidence for the ablation results in Table~\ref{tab:results_ablation_appendix}. 
The red boxes highlight representative local failures caused by removing or misplacing key components of WildIcon.
With foreground-only conditioning, the model preserves the coarse body appearance but still produces localized identity drift, such as tail deformation in later frames.
With high-frequency-only conditioning, local residual cues are available, but the lack of foreground semantic context leads to unstable appearance, including abnormal whitening and texture drift on the back.
Injecting identity tokens only into early blocks disrupts temporal formation, leading to less coherent motion and more noticeable changes in head appearance across frames.
Injecting identity conditioning into all blocks can over-constrain the generation process and introduce structural artifacts, such as distorted legs and feet.
In contrast, the full WildIcon model preserves the nyala's stripe patterns, body contour, head appearance, and limb structure more consistently across the generated frames.

\noindent\textbf{Downstream effect of identity-branch inputs.}
Figure~\ref{fig:appendix_reid_ablation_identity_branch} reports ReID gains over the real-only baseline for different identity-branch inputs.
FG-only and HF-only synthetic data both improve downstream ReID, but neither consistently matches the full WildIcon setting.
Combining foreground-isolated appearance with foreground-filtered high-frequency residuals yields the strongest gains across species and ReID backbones, indicating that the two cues are complementary for producing useful filtered synthetic training candidates.

\begin{figure*}[ht]
    \centering
    \includegraphics[width=1.0\columnwidth]{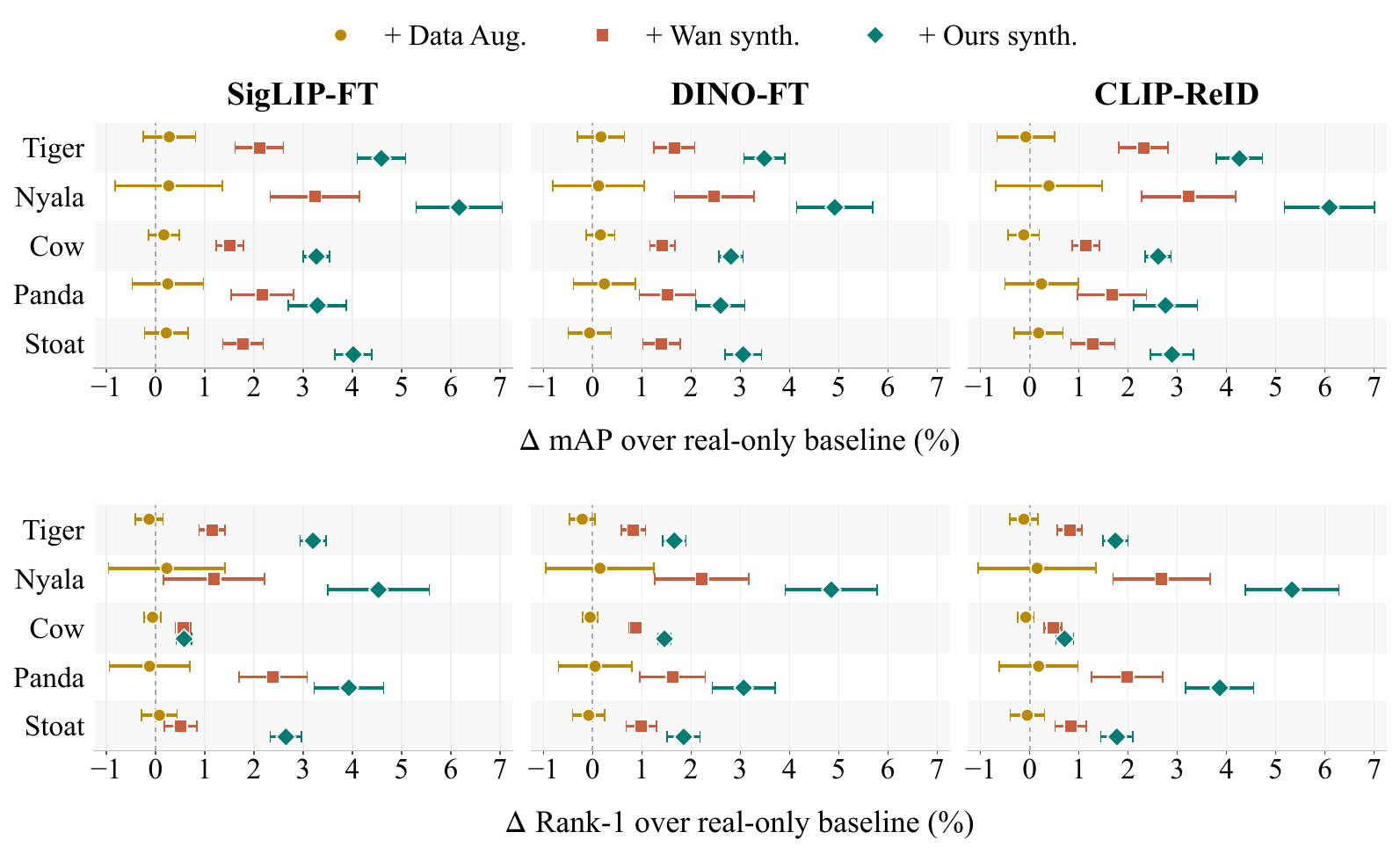}
    \caption{\textbf{Confidence intervals for downstream animal ReID gains.}
    For each backbone and species, we report the change in mAP and Rank-1 relative to the corresponding real-only baseline. Points denote mean gains over ten runs, and horizontal bars denote 95\% confidence intervals. The dashed vertical line marks zero gain.} 
    \label{fig:app_reid_results_withCI}
\end{figure*}
\subsection{ReID results with confidence interval}
Figure~\ref{fig:app_reid_results_withCI} visualizes the 95\% confidence intervals for the downstream ReID gains reported in Table~\ref{tab:animal_reid_main}. 
We plot the change in mAP and Rank-1 relative to the corresponding real-only baseline for each species and backbone. 
The dashed vertical line denotes zero gain, while points to the right indicate improved ReID performance after adding extra training samples.

Across the three backbones, classical data augmentation stays close to the zero-gain line, suggesting that simply increasing the training set through standard transformations provides limited benefit. 
In contrast, both synthetic-data settings generally improve mAP and Rank-1, with filtered WildIcon-generated samples producing the largest gains in most species-backbone combinations. 
The improvements are especially clear for mAP, where our method consistently shifts further to the right than Wan's synthetic data, indicating that the larger gains observed for WildIcon are consistent with improved reference-conditioned appearance preservation providing more useful downstream training supervision.

\subsection{Additional ReID Evaluations}
\label{app:additional_reid_evals}
\noindent\textbf{Direct generated-to-real retrieval.}
To provide a more direct evaluation of individual correspondence, we test whether generated frames can retrieve held-out real images of the source individual among same-species negatives.
For each eligible test identity, one image from the real query set is used solely as the I2V reference and excluded from the retrieval gallery.
The gallery contains other held-out real images of the same identity as positives and real gallery images of all other test identities from the same species as negatives.
The CLIP-ReID evaluator is trained only on the disjoint real training split and does not observe any test identity during training.

Using the same reference image, prompt, random seed, and generation settings, we generate one video per identity with frozen Wan2.2-A14B, Wan2.2 LoRA-fine-tuned on WildlifeVid (LoRA-FT), and WildIcon.
To reduce dependence on frames that remain very close to the reference image, we exclude frames before index 15 and evaluate the same seven predetermined positions for every method:
$\{15,25,35,45,55,65,75\}$.
No DINO-based filtering, similarity thresholding, or manual frame selection is used in this experiment.
Generated queries and real gallery images use the same animal-centric preprocessing pipeline.

As shown in Table~\ref{tab:generated_to_real}, WildIcon achieves the highest mAP and Rank-1 among the generated-query methods across all five datasets.
The real-reference queries remain stronger, as expected, while WildIcon narrows the gap more consistently than either frozen Wan2.2 or WildlifeVid LoRA fine-tuning.
This experiment provides complementary evidence that WildIcon-generated frames retain source-specific cues that support retrieval against held-out real observations.

\begin{table*}[t]
\centering
\small
\renewcommand{\arraystretch}{1.08}
\caption{\textbf{Direct generated-to-real retrieval.}
Each entry reports frame-level mAP / Rank-1 (\%).
Generated frames are used as queries against a gallery containing held-out real images of the source individual and same-species negatives.
No DINO-based filtering or manual frame selection is applied.}
\label{tab:generated_to_real}
\begin{tabular}{@{}lccccc@{}}
\toprule
Query source & Tiger & Nyala & Cow & Panda & Stoat \\
\midrule
Real reference
& 64.38 / 96.97
& 12.87 / 11.11
& 71.11 / 100.00
& 30.60 / 68.42
& 64.49 / 82.61 \\

Wan2.2-A14B
& 42.73 / 78.79
& 4.18 / 3.97
& 54.62 / 85.71
& 21.87 / 47.37
& 42.36 / 60.25 \\

LoRA-FT
& 43.91 / 82.68
& 4.95 / 5.29
& 54.74 / 88.57
& 21.68 / 49.62
& 46.83 / 67.08 \\

WildIcon
& \textbf{53.72 / 89.61}
& \textbf{10.84 / 10.32}
& \textbf{64.27 / 94.29}
& \textbf{26.14 / 58.65}
& \textbf{54.61 / 75.78} \\
\bottomrule
\end{tabular}
\end{table*}

\noindent\textbf{Sensitivity to the synthetic-sample selection representation.}
The generated-to-real experiment above is independent of DINO during both generated-frame selection and retrieval evaluation: frames are sampled at predetermined temporal positions and evaluated using CLIP-ReID.
We additionally test whether the downstream advantage depends on using DINOv3 for synthetic-sample filtering.

We replace the DINOv3 selection representation with a frozen SigLIP encoder.
Both representations operate on the same generated candidate pool and animal-centric crops.
Because their similarity scales are not directly comparable, candidates are ranked within each identity, while the same per-identity synthetic budget and per-reference cap are retained.
The ReID training protocol and real-only query/gallery sets remain unchanged, and CLIP-ReID is used as the downstream evaluator.

Table~\ref{tab:filter_robustness} reports the absolute mAP / Rank-1 difference between WildIcon and Wan2.2.
WildIcon retains positive gains across all five datasets under both DINOv3- and SigLIP-based selection, indicating that the downstream improvement is not specific to a single synthetic-sample selection representation.

\begin{table}[t]
\centering
\small
\renewcommand{\arraystretch}{1.08}
\caption{\textbf{Effect of the synthetic-sample selection representation.}
Each entry reports the absolute mAP / Rank-1 difference (\%) between WildIcon and Wan2.2 using CLIP-ReID as the downstream evaluator.}
\label{tab:filter_robustness}
\begin{tabular}{@{}lccccc@{}}
\toprule
Selection filter & Tiger & Nyala & Cow & Panda & Stoat \\
\midrule
DINOv3
& +1.95 / +0.93
& +2.86 / +2.65
& +1.47 / +0.24
& +1.09 / +1.88
& +1.61 / +0.94 \\

SigLIP
& +1.81 / +0.68
& +2.94 / +2.11
& +1.12 / +0.08
& +1.57 / +2.03
& +1.44 / +1.06 \\
\bottomrule
\end{tabular}
\end{table}

\begin{figure*}[ht]
    \centering
    \includegraphics[width=1.0\columnwidth]{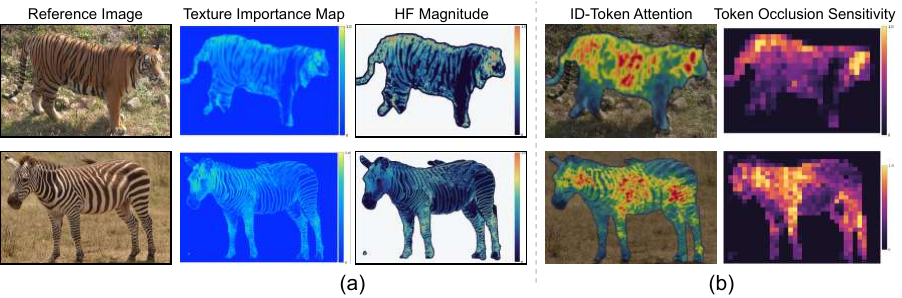}
    \caption{\textbf{Frequency-cue utilization analysis.} (a) Texture importance and high-frequency (HF) magnitude maps localize identity-related structures, such as stripe boundaries, tail patterns, and contour transitions. (b) ID-token attention and token occlusion sensitivity maps show that WildIcon attends to and depends on these high-frequency identity regions during generation.
    }
    \label{fig:frequency_cue_utilization}
\end{figure*}

\subsection{Frequency-Cue Utilization Analysis}
\label{app:frequency_cue_analysis}
To further examine whether WildIcon actively uses high-frequency identity cues during generation, we provide a frequency-cue utilization analysis in Fig.~\ref{fig:frequency_cue_utilization}. 
This analysis has two parts: cue localization and model utilization. 
The first part examines whether the high-frequency residuals spatially overlap with visible identity-discriminative regions, and the second part analyzes whether the injected identity tokens influence the denoising process.

\noindent\textbf{Cue localization.}
Fig.~\ref{fig:frequency_cue_utilization}(a) visualizes the relationship between identity-cue regions and high-frequency information. 
For each reference image, we identify identity-discriminative regions, including body stripes, facial markings, tail patterns, and contour transitions.
These regions indicate the local visual structures that are likely to distinguish one individual from another within the same species.

We visualize the foreground-filtered high-frequency residual $I_{\mathrm{hf}}$ used by WildIcon by converting it into a scalar magnitude map. 
Since $I_{\mathrm{hf}}$ is a multi-channel residual image, we average the absolute residual magnitude over color channels:
\[
H_{\mathrm{hf}}(u,v)
=
\frac{1}{C}
\sum_{c=1}^{C}
\left|
I_{\mathrm{hf}}^{(c)}(u,v)
\right|,
\]
where $(u,v)$ denotes the spatial location and $C$ is the number of image channels. 
Let $\Omega_{\mathrm{fg}}$ denote the foreground region obtained from $I_{\mathrm{seg}}$. 
For visualization, we normalize the response within $\Omega_{\mathrm{fg}}$:
\[
\widetilde{H}_{\mathrm{hf}}(u,v)
=
\frac{
H_{\mathrm{hf}}(u,v)
-
\min_{(u',v')\in\Omega_{\mathrm{fg}}} H_{\mathrm{hf}}(u',v')
}{
\max_{(u',v')\in\Omega_{\mathrm{fg}}} H_{\mathrm{hf}}(u',v')
-
\min_{(u',v')\in\Omega_{\mathrm{fg}}} H_{\mathrm{hf}}(u',v')
+
\varepsilon
}.
\]
Pixels outside the foreground region are masked out in the visualization.
Higher values in $\widetilde{H}_{\mathrm{hf}}$ indicate sharper local structures, such as stripe boundaries, texture transitions, and contour changes.
The spatial overlap between marked identity-cue regions and the normalized high-frequency response shows that the foreground-filtered high-frequency residual captures local structures relevant to individual-level discrimination, rather than high-frequency clutter from the background.

\noindent\textbf{Model utilization.}
Fig.~\ref{fig:frequency_cue_utilization}(b) further analyzes whether the generation backbone uses the high-frequency cues.
We consider two complementary visualizations: identity-token cross-attention and HF-guided identity-token occlusion contribution.

For the cross-attention analysis, we record the attention weights from video visual tokens to the unified condition sequence in the selected DiT blocks. 
Let $X_{\sigma}^{(l)}$ denote the video visual tokens at noise level $\sigma$ and selected DiT block $l\in \mathcal{S}$, where $\mathcal{S}$ is the set of middle-to-late blocks used for identity conditioning. 
The projected identity tokens are denoted by $\bar{T}_{\mathrm{id}} = P_{\mathrm{id}}(T_{\mathrm{id}})$, and the unified condition sequence is
$
T_{\mathrm{cond}}
=
[T_{\mathrm{txt}};\bar{T}_{\mathrm{id}}],
$
where $[\cdot;\cdot]$ denotes token-wise concatenation.
For attention head $h$, the query and key projections are
\[
Q_{\mathrm{vid},\sigma}^{(l,h)}
=
X_{\sigma}^{(l)} W_Q^{(l,h)},
\quad
K_{\mathrm{cond}}^{(l,h)}
=
T_{\mathrm{cond}} W_K^{(l,h)}.
\]
The cross-attention matrix is computed as
\[
A_{\mathrm{cond},\sigma}^{(l,h)}
=
\operatorname{softmax}
\left(
\frac{
Q_{\mathrm{vid},\sigma}^{(l,h)}
\left(K_{\mathrm{cond}}^{(l,h)}\right)^{\top}
}{
\sqrt{d_h}
}
\right),
\]
where $d_h$ is the feature dimension of each attention head, and the softmax is applied over the condition-token dimension.
We retain the columns corresponding to $\bar{T}_{\mathrm{id}}$, yielding the identity-attention matrix $A_{\mathrm{id},\sigma}^{(l,h)}$.
Each entry $A_{\mathrm{id},\sigma}^{(l,h)}(q,p)$ measures how strongly generated visual token $q$ attends to identity token $p$.

We aggregate the identity-attention response by averaging over selected layers, attention heads, sampled noise levels, and generated visual tokens:
\[
\alpha_p
=
\operatorname*{Mean}_{l\in \mathcal{S},\,h\in\mathcal{H},\,\sigma\in\Sigma,\,q\in\mathcal{Q}}
A_{\mathrm{id},\sigma}^{(l,h)}(q,p),
\]
where $\alpha_p$ denotes the aggregated attention score of identity token $p$.
The token-level scores $\{\alpha_p\}$ are then reshaped to the reference-image token grid and upsampled to the image resolution.
High-response regions indicate reference identity-token locations that receive stronger attention from generated visual tokens.
The resulting maps are concentrated around high-frequency identity regions, suggesting that WildIcon uses the injected identity tokens to refine local appearance cues during generation.

To provide a stronger diagnostic signal than attention alone, we also compute an HF-guided identity-token occlusion contribution map.
Let $\mathcal{G}$ denote the set of spatial identity-token groups, where each group $g\in\mathcal{G}$ corresponds to a local region in the reference image.
For each token group $g$, we construct an occluded identity-token sequence $T_{\mathrm{id}}^{\mathrm{occ}(g)}=\operatorname{Occlude}(T_{\mathrm{id}};g)$ by replacing the corresponding local identity tokens with the mean identity-token feature.
Here, $\operatorname{Occlude}(\cdot)$ keeps all identity tokens unchanged except those in group $g$, which are replaced by the mean feature computed over $T_{\mathrm{id}}$.
Let $c(T_{\mathrm{id}})$ denote the identity-augmented condition used in the selected cross-attention blocks, constructed from the text tokens and the projected identity tokens, i.e., $c(T_{\mathrm{id}})=[T_{\mathrm{txt}};P_{\mathrm{id}}(T_{\mathrm{id}})]$.
Using the same noisy latent, text prompt, and noise level, we measure the change in the predicted velocity:
\[
\Delta_g(\sigma)
=
\left\|
\Psi_{\theta}\!\left(\mathrm{z}_\sigma, \sigma, c(T_{\mathrm{id}})\right)
-
\Psi_{\theta}\!\left(\mathrm{z}_\sigma, \sigma, c(T_{\mathrm{id}}^{\setminus g})\right)
\right\|_2,
\]
where $\Psi_{\theta}$ is the velocity prediction network.
The contribution score for token group $g$ is obtained by averaging over sampled noise levels:
\[
S_g
=
\mathbb{E}_{\sigma\in\Sigma}
\left[
\Delta_g(\sigma)
\right],
\]
where $\Sigma$ denotes the set of sampled noise levels used for the analysis.
For visualization, we normalize the contribution scores across all token groups:
\[
\widetilde{S}_g
=
\frac{
S_g - \min_{g'\in\mathcal{G}} S_{g'}
}{
\max_{g'\in\mathcal{G}} S_{g'}
-
\min_{g'\in\mathcal{G}} S_{g'}
+\varepsilon
}.
\]
The normalized scores $\{\widetilde{S}_g\}$ are mapped back to the corresponding spatial token grid and visualized as a contribution heatmap.
Regions with higher contributions indicate that occluding the corresponding local identity-token group causes a larger change in the denoising prediction.
Compared with cross-attention, this perturbation-based visualization provides a more direct diagnostic of which HF-guided identity regions influence the generation dynamics.

\begin{table*}[ht]
\centering
\renewcommand{\arraystretch}{1.08}
\caption{\textbf{Efficiency and memory profile.}
We report 81-frame training and inference costs for Wan2.2-I2V-A14B and WildIcon at $832\times480$ resolution.
Training time is reported as hours per 1K training samples. Memory is reported as peak per-GPU memory in GiB.}
\label{tab:efficiency}
\begin{tabular}{@{}llcccc@{}}
\toprule
Method
& GPUs
& \makecell{Train\\h/1K samples}
& \makecell{Infer.\\s/video}
& \makecell{Train mem.}
& \makecell{Infer. mem.} \\
\midrule
\multirow{2}{*}{Wan2.2}
& 1 GPU
& --
& 693.3
& --
& 51.36 \\
& 4 GPUs
& --
& 300.5
& --
& 45.03 \\
\midrule
\multirow{2}{*}{WildIcon}
& 1 GPU
& 5.2
& 701.5
& 51.67 
& 55.73 \\
& 4 GPUs
& 1.3
& 315.8
& 54.48 
& 46.47 \\
\bottomrule
\end{tabular}
\end{table*}

\subsection{Efficiency and Cost Analysis}
Table~\ref{tab:efficiency} reports the computational cost and peak GPU memory of WildIcon.
All inference results are for 81-frame videos at $832\times480$ resolution with 50 sampling steps.
Compared with the frozen Wan2.2-A14B backbone, WildIcon introduces only a small inference overhead.
On a single GPU, inference time increases from 693.3s to 701.5s per video, and peak inference memory increases from 51.36 GiB to 55.73 GiB.
Under the 4-GPU inference configuration, inference time increases from 300.5s to 315.8s per video, with peak per-GPU memory increasing from 45.03 GiB to 46.47 GiB.

For training, WildIcon requires 5.2 hours per 1K training samples on a single GPU and 1.3 hours per 1K training samples under the 4-GPU configuration.
The peak training memory remains below 55 GiB per GPU in both settings, fitting comfortably within the 96 GiB memory budget of NVIDIA RTX PRO 6000 Blackwell GPUs.
These results indicate that WildIcon adds a manageable training cost and modest inference overhead while preserving the practical efficiency of the frozen Wan2.2-A14B backbone.

\section{WildlifeVid: Additional Details}
\label{app:wildvid_details}

\begin{table*}[ht]
\centering
\caption{Comparison of existing domain-specific video generation datasets and our WildlifeVid.}
\setlength{\tabcolsep}{3pt}
\label{tab:dataset_comparison}
\begin{tabular}{@{}lcccccc@{}}
\toprule
Dataset & Year & Domain & \# Videos & Avg. Len (s) & Duration (h) & Resolution \\ \midrule
UCF-101~\cite{soomro2012ucf101} & 2012 & Human (Action) & 13.3K & 7.2 & 26.7 & 240p \\
NTU RGB+D~\cite{shahroudy2016ntu} & 2014 & Human (Action) & 114K & -- & 37 & 1080p \\
MSP-Avatar~\cite{sadoughi2015msp} & 2015 & Human (Action) & 74 & -- & 3 & 1080p \\
Taichi-HD~\cite{siarohin2019first} & 2019 & Human (Action) & 3K & -- & -- & 256p \\
SkyTimelapse~\cite{xiong2018learning} & 2018 & Sky & 35K & -- & -- & 360p \\
FaceForensics++~\cite{rossler2019faceforensics++} & 2019 & Human (Face) & 1K & -- & -- & Diverse \\
TikTok-v4~\cite{chang2024magicpose} & 2023 & Human (Dance) & 350 & -- & 1 & -- \\
ConsIDVid\cite{wu2026considgen} & 2025 & Rigid Object & 44.3K & 8.4 & 104 & Diverse \\ \midrule
WildlifeVid (Ours) & 2026 & Wildlife & 37K & 9.3 & 96.4 & Diverse \\ \bottomrule
\end{tabular}
\end{table*}
\subsection{Comparison with Existing Video Datasets}
\label{app:wildvid_comparison}
Recent efforts~\cite{chen2024panda, zhao2026unlearning,pang2025cabin,wang2025koala,qiao2026multiple} in video generation primarily focus on collecting large, general-purpose video datasets to train video generative models. 
However, domain-specific video datasets remain limited in both scale and diversity. 
As shown in Table~\ref{tab:dataset_comparison}, many existing domain-focused resources are mostly human-centric (\textit{e.g.}, UCF-101~\cite{soomro2012ucf101}, Taichi-HD~\cite{siarohin2019first}, FaceForensics++~\cite{rossler2019faceforensics++}), making them inadequate for capturing fine-grained wildlife identity. 

\subsection{Video Collection}
\label{app:wildvid_sources}
WildlifeVid is constructed by curating and re-annotating clips from four publicly accessible wildlife video sources, including Animal-in-Motion (AiM)~\citep{zhao2025web}, AnimalKingdom~\citep{Ng_2022_CVPR}, MammalNet~\citep{chen2023mammalnet}, and LoTE-Animal~\citep{Liu_2023_ICCV}.
These sources provide broad coverage of wildlife appearance, behavior, viewpoint, and environmental variation, but their original annotations are not sufficient for individual-consistent I2V training and evalutation. 
In particular, they are designed for tasks such as animal reconstruction, species recognition, action understanding, behavior analysis, and pose estimation, rather than individual-identity modeling. 
We therefore use these datasets as candidate video sources and apply a unified curation and re-annotation pipeline to obtain training-ready WildlifeVid clips.

AiM is a web-scale resource for 4D animal reconstruction, converting videos into object-centric clips with auxiliary annotations, such as masks, keypoints, and depth. 
The AiM benchmark contains 30K videos and 2M frames. 
AnimalKingdom is a large and diverse benchmark spanning 850 species across six major animal classes, with annotations for video grounding, fine-grained action recognition, and pose estimation. 
MammalNet contains over 18K videos totaling 539 hours, covering 173 mammal categories and 12 high-level behaviors; and 
LoTE-Animal focuses on understanding the behavior of endangered animals, with 10K annotated video sequences and 28K images collected over 12 years from 11 endangered species.

\begin{figure*}[t]
    \centering
    \includegraphics[width=\textwidth]{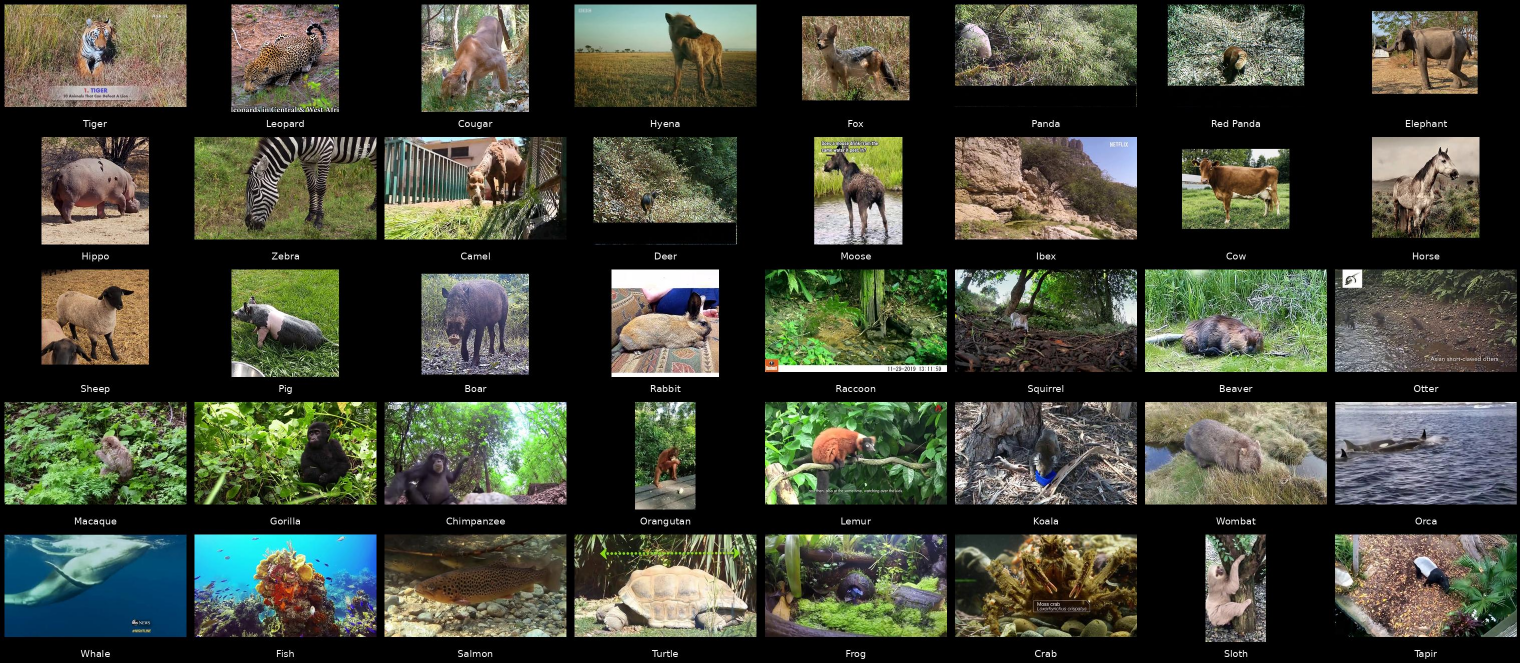}
    \caption{Random examples from WildlifeVid.}
    \label{fig:wildlifevid_reference_gallery}
\end{figure*}

Each retained WildlifeVid sample is represented as a tuple:
\[
(\mathcal{C}, y_{\mathcal{C}}, a_{\mathcal{C}}, \mathrm{id}_{\mathcal{C}}),
\]
where $\mathcal{C}$ denotes a retained video clip, $y_{\mathcal{C}}$ is its standardized text prompt, $a_{\mathcal{C}}$ denotes the species label, and $\mathrm{id}_{\mathcal{C}}$ denotes the clip-level individual identity label.
Figure~\ref{fig:wildlifevid_reference_gallery} presents some random examples from WildlifeVid.

\subsection{Licenses}
\label{appendix:licenses}
WildlifeVid is constructed by curating and re-annotating clips from four existing wildlife video resources: Animal-in-Motion (AiM), AnimalKingdom, MammalNet, and LoTE-Animal. 
We cite the original creators of all source datasets and follow their corresponding licenses, access conditions, and terms of use. 
The upstream dataset licenses and access terms are summarized below.

\begin{itemize}
    \item \textbf{Animal-in-Motion (AiM)}~\citep{zhao2025web}: 
    The official repository is released under the Apache-2.0 License and is available at the 
    \href{https://github.com/briannlongzhao/Animal-in-Motion}{official repository}. 
    
    \item \textbf{MammalNet}~\citep{chen2023mammalnet}: 
    The dataset annotations are distributed under CC BY, and the official code repository is released under the MIT License. 
    The official repository is available at the 
    \href{https://github.com/Vision-CAIR/MammalNet}{MammalNet repository}. 

    \item \textbf{AnimalKingdom}~\citep{Ng_2022_CVPR}:  
    The dataset is available through the 
    \href{https://sutdcv.github.io/Animal-Kingdom/}{official project page}. 
    We submitted the 
    \href{https://docs.google.com/forms/d/e/1FAIpQLSccqs92uqbaf1jfHrWB7lMejCDp8galelSRhp4u9DpX5-x3YQ/viewform}{official download request form} 
    and obtained permission from the authors.

    \item \textbf{LoTE-Animal}~\citep{Liu_2023_ICCV}: 
    The dataset is released under CC BY-SA 4.0 and is available through the 
    \href{https://lote-animal.github.io/}{official project page}. 
\end{itemize}

\subsection{Data Curation Pipeline: Implementation Details}
\label{app:wildvid_pipeline}

\noindent\textbf{Shot splitting.}
Raw wildlife videos may contain shot changes, abrupt camera transitions, or multiple temporal segments with different visible individuals.
We first split each source video into candidate clips with PySceneDetect~\citep{pyscenedetect}, which detects large values of weighted average pixel change across frames.
To avoid over-segmentation caused by brief visual fluctuations, we merge very short adjacent segments during shot splitting.
This setting is used only to stabilize shot-boundary detection and does not impose a fixed duration threshold on the final retained clips.
After this step, each candidate clip is further checked for species presence, single-subject visibility, and temporal identity consistency.

\noindent\textbf{CLIP-based species filtering.}
We remove candidate clips that do not clearly contain the target species.
For a candidate clip $\mathcal{C}$ with $L_{\mathcal{C}}$ decoded frames, we uniformly sample $n_{\mathcal{C}}$ frames over its duration and compare them with a species-level text prompt using CLIP~\citep{radford2021learning,hessel2021clipscore}.
Let $J_i$ denote the $i$-th sampled frame, and let $\mathrm{CLIP}_{\mathrm{img}}(\cdot)$ and $\mathrm{CLIP}_{\mathrm{text}}(\cdot)$ denote the CLIP image and text encoders, respectively.
Given the source species label $a_{\mathcal{C}}$, we compute
\[
s_{\mathrm{sp}}(\mathcal{C})
=
\frac{1}{n_{\mathcal{C}}}
\sum_{i=1}^{n_{\mathcal{C}}}
\operatorname{sim}\!\left(
\mathrm{CLIP}_{\mathrm{img}}(J_i),
\mathrm{CLIP}_{\mathrm{text}}(\text{``a photo of } a_{\mathcal{C}}\text{''})
\right).
\]
Candidate clips with $s_{\mathrm{sp}}(\mathcal{C}) < \rho_{\mathrm{sp}}$ are discarded.
In our implementation, we set $\rho_{\mathrm{sp}}=0.25$.

\noindent\textbf{Single-subject enforcement.}
Since WildlifeVid is designed for individual-consistent I2V generation, each retained clip should contain one dominant wildlife individual without identity changes.
To enforce this constraint, we use a Grounded-SAM2 tracking pipeline, where GroundingDINO-Tiny initializes text-grounded wildlife detections and SAM2 Hiera Large predicts and propagates instance masks over time.
Bounding boxes are derived from the predicted instance masks. 
In our implementation, videos are resampled to $10$ fps, new text-grounded detections are introduced every $50$ frames, and the detection thresholds are set to box\_threshold=0.6 and text\_threshold=0.3.
Very small detections are removed based on the mask-derived bounding-box area to suppress unreliable distant or spurious subjects.
A candidate clip is rejected if any checked frame contains no valid wildlife instance or more than one valid wildlife instance.

\noindent\textbf{Visibility and temporal stability filtering.}
After single-subject enforcement, we further refine the remaining clips to ensure reliable identity supervision.
To remove unstable tracks, we compute the Intersection over Union (IoU) between bounding boxes in adjacent checked frames.
If the IoU of any adjacent checked-frame pair falls below $0.5$, the whole track is removed, since such cases usually indicate missed detections, tracking failure, abrupt shot changes, or possible identity switches.
We also remove tracks in which the animal is heavily truncated by the image boundary.
Specifically, if the mask-derived bounding box lies within $3$ pixels of any image boundary in any checked frame, the corresponding track is discarded.

\begin{figure*}[t]
\centering
\begin{tcolorbox}[
    colback=gray!3!white,
    colframe=gray!60!black,
    colbacktitle=gray!15!white,
    coltitle=black,
    fonttitle=\bfseries,
    boxrule=0.6pt,
    arc=2mm,
    left=2mm,
    right=2mm,
    top=1mm,
    bottom=1mm,
    width=\textwidth,
    sharp corners,
    title=\textbf{Instruction Template for Initial Wildlife Prompt Generation},
]
\textbf{System Prompt:} \\
You are an expert wildlife video captioner.  
Your task is to write a short prompt-style description of a wildlife video clip using only directly visible content.

\medskip
\textbf{Rules:}
\begin{itemize}[leftmargin=1.5em, topsep=2pt, itemsep=2pt]
    \item Write exactly 1–2 sentences under 40 words.
    \item Start with the animal species if identifiable.
    \item Describe visible motion, pose, or viewpoint.
    \item Optionally include brief scene context if clearly visible.
    \item Use concise prompt-style language rather than full narrative language.
    \item Do not mention camera movement, video quality, or artistic style.
    \item Do not speculate about identity, intent, emotion, or invisible attributes.
    \item Output only the prompt text.
\end{itemize}

\medskip
\textbf{User Prompt:} \\
Write one short prompt-style sentence describing the wildlife clip using only visible content.  
Include the species, visible motion or pose, viewpoint if clear, and brief scene context if relevant.  
Do not mention camera behavior or make any unsupported inference.
\end{tcolorbox}
\caption{Instruction template for initial wildlife caption generation.}
\label{fig:llm_prompt_template}
\end{figure*}

\noindent\textbf{Prompt generation and cleaning.}
Text prompts provide controllable conditions for WildlifeVid clips, including the visible species, motion, viewpoint, and scene context.
For each retained clip, we sample representative frames over its duration and query GPT-5~\citep{openai2025gpt5api} with an instruction to describe only visible content in the sampled frames.
The instruction template used for this step is shown in Figure~\ref{fig:llm_prompt_template}.
Since raw captions may contain generic template phrases, unsupported behavioral interpretations, repeated descriptions, or details unrelated to the target animal, we further clean them into standardized training prompts.
Specifically, we remove non-visual or speculative content, such as inferred intent, emotion, causal explanations, or attributes that are not clearly visible in the sampled frames.
We also remove irrelevant background descriptions when they do not affect the target animal or the requested motion.
The final prompt is a concise factual sentence describing the visible animal, its motion, viewpoint, and scene context, \textit{e.g.}, ``a tiger walks slowly through grass in a side view''.

\noindent\textbf{Identity Annotation Details}
WildlifeVid assigns identity labels at the clip level.
After curation, each retained track is first treated as a candidate identity unit.
To associate clips that contain the same individual, we perform within-species candidate grouping using visual embeddings.
For each retained clip, we use the temporal midpoint frame as the representative image and extract an appearance embedding with a frozen DINOv3 ViT-H/16+ model~\citep{simeoni2025dinov3}.
All embeddings are $\ell_2$-normalized before similarity computation.
Within each species, we perform average-linkage agglomerative clustering in cosine space with a distance threshold of $0.15$, corresponding approximately to candidate cross-clip merges with cosine similarity at least $0.85$.

High-similarity candidate merges are then manually reviewed.
Annotators compare stable individual-specific appearance cues, including stripe topology, spot configuration, facial patches, horn shape, tail markings, contour details, and other persistent coat-pattern cues when applicable.
A merge is accepted only when all the annotators agree that the clips contain the same individual; otherwise, the candidate is rejected.
Accepted clips are assigned the same identity label, while rejected candidates retain their own identities.

\subsection{Benchmark Protocol and Data Splits}
\label{app:wildlifevid_benchmark}
We evaluate all models on a fixed wildlife I2V benchmark constructed from WildlifeVid.
Benchmark identities are drawn from all four source datasets used to build WildlifeVid~\citep{zhao2025web,Ng_2022_CVPR,chen2023mammalnet,Liu_2023_ICCV}. 
The benchmark contains 10 species and 100 identities in total, with 10 identities per species. 
For every identity, we select one reference image and pair it with two prompts, resulting in 200 generation cases overall.
Thus, each generation case is defined as a tuple of \emph{(species, identity, reference image, prompt)}.

\noindent\textbf{Prompt design.}
Each identity is paired with two prompts to evaluate identity preservation under different levels of motion and appearance variation.
Prompt A contains mild motion and limited viewpoint change, such as slow forward movement, gentle body motion, or a slight head turn.
Prompt B is more challenging and introduces stronger pose or viewpoint variation, such as turning around, body rotation, or larger camera movement.

\noindent\textbf{Data splits and fixed evaluation protocol.}
The benchmark is kept fixed throughout all model comparisons and ablation studies. 
To avoid identity leakage between training and evaluation, all benchmark identities are excluded from the WildlifeVid training split, and the model is trained only on the remaining identities. 
All compared methods are evaluated using the same reference images, prompts, output resolution, and video duration.

\section{Limitation}
\label{app:limitation}
This work focuses on a controlled, individual-consistent I2V setting in which the reference image contains a known wildlife individual, and the generated video is expected to preserve that individual's appearance across new poses, viewpoints, and motions. 
Consistent with this formulation, WildlifeVid prioritizes single-subject, temporally coherent clips so that identity labels, prompts, and evaluation targets are well defined. 
This design makes the benchmark suitable for studying individual-level identity preservation. In contrast, broader full-scene wildlife generation involving multiple interacting animals, dense occlusion, or ambiguous identity association remains a complementary direction.

WildIcon operates in a single-reference setting and therefore cannot create new ground-truth identity evidence beyond what is supported by the conditioning image.
When generation requires substantial viewpoint changes, particularly exposure of identity-bearing regions that are not visible in the reference, the model may extrapolate or hallucinate local markings, textures, or contours.
Accordingly, individual consistency in this work refers to maintaining identity-relevant appearance supported by the reference image, rather than recovering verified appearance on unseen surfaces.
Generated videos should therefore be treated as synthetic training candidates, not as genuine wildlife observations, ecological records, retrieval-database entries, or substitutes for real identity evidence.

WildIcon also depends on the visual evidence available in the reference image. 
The method is applicable when individual-discriminative cues, such as fur texture, stripe boundaries, spot configurations, or contour transitions, are visible in the input. 
When the reference image is extremely low-resolution, strongly blurred, heavily occluded, captured under very poor lighting, or depicts species with weak external individual markings, the available identity-conditioning signal is inherently limited. 

\section{Broader Societal Impacts}
\label{appendix:broader_impacts}
\noindent\textbf{Societal Impacts.}
WildIcon and WildlifeVid are developed to support data-scarce wildlife identification.
The potential positive impact of WildIcon lies in using filtered generated samples as candidate training augmentations when real observations are limited.
These samples can expose ReID models to additional appearance and motion variation during training, but they do not constitute additional field observations or new ground-truth identity evidence.
Improved ReID models may in turn support wildlife monitoring workflows, while ecological inference and validation should continue to rely on real observations.

At the same time, the proposed technology may have unintended negative impacts. 
First, synthetic wildlife videos could be misused to fabricate evidence of animal sightings, population presence, or conservation outcomes. 
Such misuse may be especially harmful when generated media are presented without clear disclosure. 
Second, because the generated videos are synthetic observations rather than field-captured evidence, they should not be used directly to support ecological claims, such as confirming the presence, absence, movement, or population size of a species. 
Their intended role is to support model training and controlled evaluation, not to replace real observations or expert ecological assessment.

\noindent\textbf{Safeguards and Responsible Release.}
We take several safeguards for responsible release. 
The dataset, WildlifeVid, is constructed from publicly accessible wildlife video sources and further curated into single-subject, temporally stable clips with prompts and individual-level identity labels. 
We release only the curated wildlife-centric data and associated annotations permitted by the corresponding source terms. 
We avoid releasing sensitive ecological metadata, such as precise camera locations, deployment identifiers, or fine-grained timestamps, when such information could increase the risk of unauthorized tracking or disturbance of wildlife.

For generated content, all synthetic videos released with the paper are labeled as synthetic and are intended for research, model evaluation, and downstream ReID training rather than as ecological evidence. 
We do not encourage the use of generated videos to support claims about real animal sightings, population presence, or conservation outcomes. 
Any released model weights, code, or generation examples will be accompanied by usage guidelines that prohibit presenting synthetic wildlife videos as real camera-trap footage or using the model to fabricate ecological evidence.



\end{document}